\documentclass[sigconf,authordraft]{acmart}

\usepackage{multirow}
\usepackage{multicol}
\usepackage{subcaption}
\usepackage{lscape}
\usepackage{todonotes}
\usepackage{pifont}

\AtBeginDocument{%
  \providecommand\BibTeX{{%
    \normalfont B\kern-0.5em{\scshape i\kern-0.25em b}\kern-0.8em\TeX}}}

\setcopyright{acmlicensed}
\copyrightyear{2018}
\acmYear{2018}
\acmDOI{XXXXXXX.XXXXXXX}

\begin{document}

\title{Multi-Task Learning for Routing Problem with \\Cross-Problem Zero-Shot Generalization}

\author{Fei Liu}
\email{fliu36-c@my.cityu.edu.hk}
\affiliation{%
  \institution{City University of Hong Kong}
  \country{Hong Kong SAR}
}

\author{Xi Lin}
\email{xi.lin@my.cityu.edu.hk}
\affiliation{%
  \institution{City University of Hong Kong}
  \country{Hong Kong SAR}
}

\author{Zhenkun Wang}
\email{wangzk3@sustech.edu.cn}
\affiliation{%
  \institution{Southern University of Science and Technology}
  \country{Shenzhen, China}
}

\author{Qingfu Zhang}
\email{qingfu.zhang@cityu.edu.hk}
\affiliation{%
  \institution{City University of Hong Kong}
  \country{Hong Kong SAR}
}
\author{Xialiang Tong}
\email{tongxialiang@huawei.com}
\affiliation{%
  \institution{Huawei Noah’s Ark Lab}
  \city{Shenzhen}
  \country{China}
}

\author{Mingxuan Yuan}
\email{yuan.mingxuan@huawei.com}
\affiliation{%
  \institution{Huawei Noah’s Ark Lab}
  \country{Hong Kong SAR}
}

\settopmatter{printacmref=false}
\setcopyright{none}
\renewcommand\footnotetextcopyrightpermission[1]{}
\pagestyle{plain}


\begin{abstract}

Vehicle routing problems (VRPs), which can be found in numerous real-world applications, have been an important research topic for several decades. Recently, the neural combinatorial optimization (NCO) approach that leverages a learning-based model to solve VRPs without manual algorithm design has gained substantial attention. However, current NCO methods typically require building one model for each routing problem, which significantly hinders their practical application for real-world industry problems with diverse attributes. In this work, we make the first attempt to tackle the crucial challenge of cross-problem generalization. In particular, we formulate VRPs as different combinations of a set of shared underlying attributes and solve them simultaneously via a single model through attribute composition. In this way, our proposed model can successfully solve VRPs with unseen attribute combinations in a zero-shot generalization manner. Extensive experiments are conducted on eleven VRP variants, benchmark datasets, and industry logistic scenarios. The results show that the unified model demonstrates superior performance in the eleven VRPs, reducing the average gap to around 5\% from over 20\% in the existing approach and achieving a significant performance boost on benchmark datasets as well as a real-world logistics application. The source code is included in https://github.com/FeiLiu36/MTNCO.

\end{abstract}


%




\maketitle


\section{Introduction}

The vehicle routing problem (VRP) is a crucial topic in combinatorial optimization and operation research. It is widely studied in academia and has significant practical importance in real-world applications such as logistics, transportation, retail distribution, waste collection, and manufacturing~\citep{toth2014vehicle}. The objective of the VRP is to optimally manage a fleet of vehicles, minimizing the total cost while satisfying the demands of customers. Real-world industry routing problems have diverse attributes (e.g., the capacities of vehicles, the time window constraints of requests), which result in numerous VRP variants~\citep{braekers2016vehicle,vidal2020concise}. Developing a separate algorithm for each VRP is very costly and impractical. Therefore, it is desirable to build a single unified solver for solving VRPs, which can significantly reduce the overall management cost and improve operational efficiency for companies. A few attempts have been made to develop a unified heuristics~\citep{vidal2013hybrid,rabbouch2021efficient,errami2023vrpsolvereasy}, but they demand much algorithm design effort with domain knowledge from experts.


Neural combinatorial optimization (NCO) learns a neural network based heuristic to solve combinatorial optimization problems. This approach has received growing research attention due to its potential ability to generate high-quality solutions without much human effort~\citep{bengio2021machine,vinyals2015pointer,kool2018attention}. However, existing NCO methods work in a single-task manner~\citep{li2022overview,bai2023analytics}. In other words, they need to train a neural network model for each optimization problem. When the problem changes, another model must be trained from scratch, which inevitably leads to high computational costs. Some attempts to overcome this shortcoming include transfer learning and multiobjective learning~\citep{feng2020towards,li2021deep, Zhang2022meta,lin2022pareto}. However, the neural network models generated by these works can only be used to solve problems whose instances have been used for training. In other words, generalization across different problems has not been well addressed.




This paper proposes a multi-task learning approach for cross-problem generalization on vehicle routing problems to tackle the challenge. We develop a unified neural network model with attribute composition to handle multiple VRPs which can be efficiently trained by reinforcement learning (RL) without any labeled solution. We show that the unified model can be used to solve VRP variants in a zero-shot manner significantly outperforming the existing approach. Our contributions are summarized as follows: 

\begin{itemize} 
    \vspace{-2pt}
    \item We propose a novel learning-based method to tackle cross-problem generalization in VRPs. It treats the VRP variants as different combinations of a set of shared underlying attributes, and solves various VRPs simultaneously in an end-to-end multi-task learning manner. To the best of our knowledge, this is the first work to investigate cross-problem neural solvers for VRPs.
    \vspace{4pt}

    

    \item We develop a unified attention model with an attribute composition block. The proposed model has a promising zero-shot generalization ability to handle any combination of the basic attributes. 
 \vspace{4pt}
    \item We achieve state-of-the-art (SOTA) cross-problem generalization performance on eleven routing problems. This results in reducing the average gap to around 5\% from over 20\% in the existing approach. Further validation of our zero-shot generalization approach on real-world benchmark datasets and an industry logistic application demonstrates a notable performance boost.

    

\end{itemize}

\section{Related Work}

\subsection{Neural Combinatorial Optimization (NCO)} 
NCO~\citep{bengio2021machine,vinyals2015pointer,kool2018attention} intends to automatically learn a neural network-based heuristic to solve the combinatorial optimization problem. Compared to the other approaches (e.g., exact methods and heuristics), it requires very little domain-specific knowledge and usually generates high-quality solutions significantly fast. As a result, it has gained much attention in the past decade~\citep{bengio2021machine}. 

There are mainly two groups of works along this line: end-to-end methods~\citep{vinyals2015pointer, bello2016neural, nazari2018reinforcement, kwon2020pomo,joshi2022learning,choo2022simulation,pan2023h-tsp} and improvement-based methods~\citep{chen2019learning,hottung2019neural,chen2019learning,kool2022DPDP}. The former aims to construct a solution without any assistance from non-learning methods, while the latter incorporates additional algorithms to improve performance. In this paper, we focus on the end-to-end approach.

\subsection{NCO for Vehicle Routing Problem (VRP)}
NCO has been successfully applied to many vehicle routing problems, including traveling salesman problem (TSP)~\citep{bello2016neural}, capacitated VRP (CVRP)~\citep{nazari2018reinforcement}, VRP with time windows (VRPTW)~\citep{zhao2020hybrid}, open VRP (OVRP)~\citep{tyasnurita2017learning}, VRP with pickup and delivery~\citep{li2021heterogeneous}, and heterogeneous VRP~\citep{li2021deep}. A recent survey of the works on learning-based methods for different vehicle routing problems can be found in~\citet{li2022overview}. 

Despite extensive studies, the existing works have been conducted in a single-task manner, in which an individual neural model is trained for each problem. The time-consuming training process for every new problem hinders their practical application. It should be noted that various VRPs have common underlying attributes. Nevertheless, these similarities and correlations have not been adequately studied in the context of NCO. Recently, \citet{jiang2022learning,bi2022learning} and \citet{geisler2022generalization} explored robust optimization over multiple distributions. Several works~\citep{fu2021generalize,pan2023h-tsp,manchanda2023generalization,cheng2023select,drakulic2023bq,gao2023generalizable} studied generalization to large-scale problems, and \citet{zhou2023towards} considered generalization in terms of both problem size and distribution. However, the cross-problem generalization has not been studied for routing problems.

\subsection{Multi-Task Learning and Zero-Shot Learning} 
Multi-task learning (MTL) tackles multiple related learning tasks in a single learning process. It has been widely studied in various research fields, including computer vision~\citep{yuan2012visual}, bioinformatics~\citep{he2016novel}, and natural language processing~\citep{collobert2008unified}. However, MTL has received limited attention on combinatorial optimization problems. \citet{reed2022generalist} and \citet{ibarz2022generalist} proposed a general agent capable of solving diverse tasks, including several combinatorial optimization problems. \citet{wang2023efficient} presented a multi-task learning method for combinatorial optimization problems with separate encoders and decoders. Nevertheless, their approach falls short with respect to handling complicated VRPs, and they require revision or fine-tuning to solve new problems that were not previously encountered during training. 

Zero-shot learning (ZSL) allows the recognition of previously unseen objects based on their shared semantic properties or attributes~\citep{oh2017zero,xian2018zero,ruis2021independent}. Our idea of learning on multiple VRPs with several underlying attributes and generalizing to unseen VRPs is similar to compositional ZSL~\citep{ruis2021independent}, which composes novel problems out of known subparts or attributes. In the context of routing problems, the attributes introduce additional constraints to the solution generation process, distinguishing them from the features or labels investigated in computer vision tasks~\citep{ruis2021independent}.

\section{Problem Statement and Motivation}
In this section, we begin by introducing the basic CVRP formulation and then demonstrate that other VRP variants can be considered as extensions of the basic CVRP by incorporating additional attributes. We denote a CVRP on an undirected graph $G = (V, E)$. $V=\{v_0,\dots,v_n\}$, where $v_0$ is the depot and $v_1,\dots,v_n$ are $n$ customers. $V_c=\{v_1,\dots,v_n\}$ is the customer set. For the $i$-th customer, there is a demand $d_i$. $E=\{e_{ij}\},i,j\in \{1,\dots,n\}$ are the edges between every two nodes. For each edge $e_{ij}$, there is an associated cost (distance) $c_{ij}$. A fleet of homogeneous vehicles with a capacity of $C$ is sent out from the depot to visit the customers and return to the depot. Each customer must be visited once. The objective is to minimize the total travel distance of all used vehicles. 





\begin{figure*}[t]
    \centering
    \includegraphics[width=0.8\linewidth]{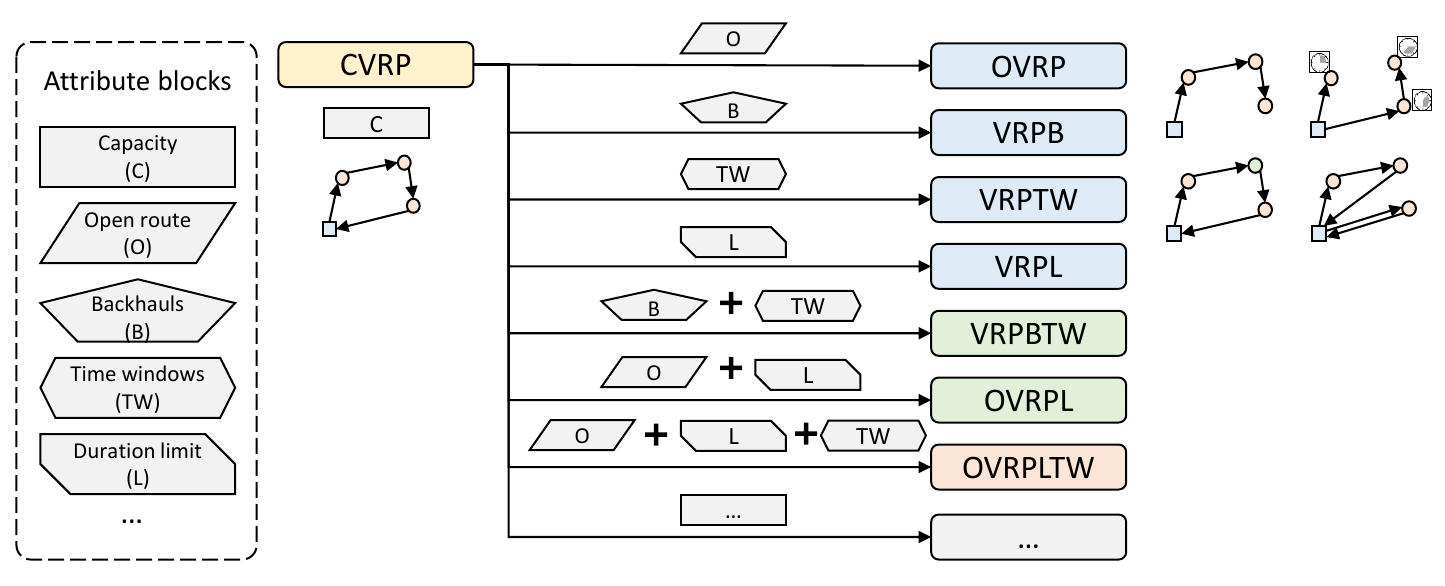}
    \caption{VRP variants as combinations of attribute blocks. The basic version is known as the Capacitated Vehicle Routing Problem (CVRP). VRP variants can be regarded as extensions of CVRP, encompassing additional attributes. For example, VRPTW extends CVRP by incorporating time windows, while OVRPTW adds an open routes attribute alongside time windows. 
    }
    \label{fig:vrps}
\end{figure*}

Figure~\ref{fig:vrps} shows that various VRPs can be regarded as extensions of CVRP by considering one or more attributes. For example, VRPTW is extended from CVRP by adding time windows, and OVRPTW involves both time windows and open route attributes.

Except for the capacity constraints (C), we involve the following attributes in this paper: 
\begin{itemize} 
    \item \textbf{Time Windows (TW)}: we denote the time window for the $i$-th node as $[e_i,l_i],i \in \{0,\dots,n\}$, where $e_i$ and $l_i$ are the early and late time windows. In addition, each node has a service time $s_i$. We consider hard time windows, i.e., the vehicle must visit the node $i$ in the time range from $e_i$ to $l_i$. If the vehicle arrives at node $i$ earlier than $e_i$, the vehicle has to wait until $e_i$.
    
    \item \textbf{Open Routes (O)}: open routes mean that the vehicle does not need to return to the depot after it services all the customers on its route.
    
    \item \textbf{Backhauls (B):} the classic CVRP assumes that all vehicles load demands at the depot and unload at customers. We call these customers, who require deliveries $d_i>0$, linehaul customers. Correspondingly, backhaul customers are those customers that need pickup $d_i<0$. We consider the VRP with mixed linehaul and backhaul customers, i.e., the order of linehaul and backhaul customers can be mixed up in each route. 
    
    \item \textbf{Duration limits (L):} duration limits refer to the situation in which the total length of the routes cannot exceed some pre-set thresholds. It is commonly used in real-world application scenarios to maintain a reasonable workload for different routes. In our setting, we use the same duration limit for all routes.
    
\end{itemize}


\textbf{Motivation}
In real-world industrial applications, there is a crucial demand to use a single model to solve various VRPs with different attributes. However, all current NCO methods need to build separate independent models to tackle these routing problems. Unlike previous approaches, this work treats the VRP variants as extensions of the basic CVRP with different combinations of the basic attributes (e.g., TW, O, B, and L). By leveraging the similarities and correlations among VRPs with shared underlying attributes, in this work, we propose to build the first cross-problem neural solver to solve various VRPs via a single model in a zero-shot generalization manner.


\section{Attribute-Sharing Model}



\begin{figure*}[t]
    \centering
    \includegraphics[width=0.9\linewidth]{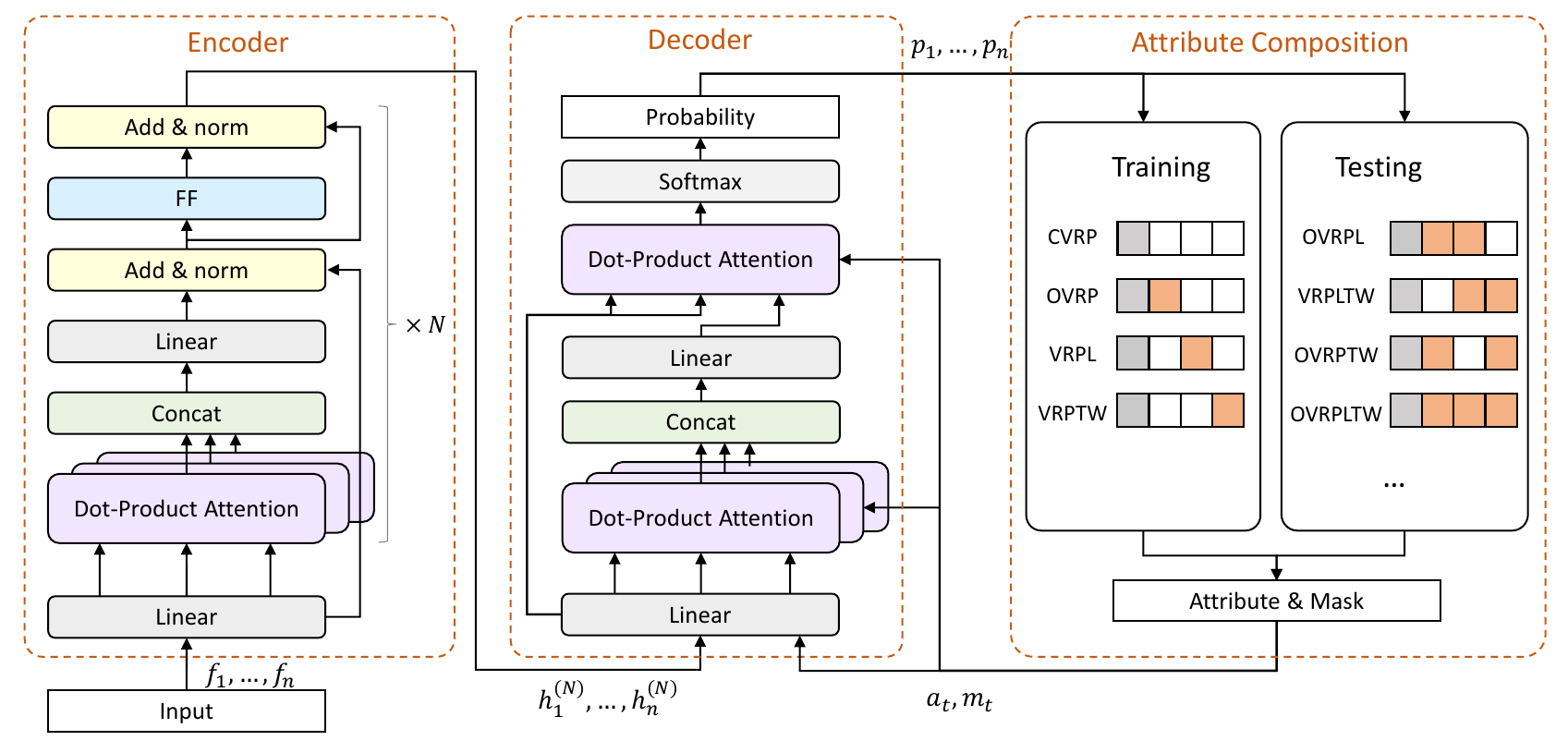}
    \caption{Unified model extended from attention model. The model is trained on multiple VRPs with diverse attributes. Then it can be used to solve numerous unseen VRPs as any combinations of the attributes involved in the training. }
    \label{fig:model}
\end{figure*}

We consider multiple VRPs as a set of related tasks and propose training a unified neural model through reinforcement learning to solve them simultaneously. Figure~\ref{fig:model} illustrates the unified model used in this work. It consists of three parts: encoder, decoder, and attribute composition. We adopt the typical encoder-decoder framework of the Attention Model (AM)~\citep{kool2018attention}. The encoder learns the node embeddings, and the decoder generates solutions sequentially. Different from the existing works~\citep{kool2018attention,kwon2020pomo,zhu2023accelerated}, we enable its ability to handle various VRPs by adding an additional attribute composition block. The idea is that the diverse VRPs actually consist of several common underlying attributes. By learning from these attributes, we can solve an exponential number of new VRPs as any combination of them.

\subsection{Model Structure}


\textbf{Encoder} 
The encoder consists of $N$ stacked multi-head attention (MHA) blocks~\citep{kool2018attention}. The input of the encoder is the node features $f_i,i=1,\dots,n$. In this paper, the input features for the $i$-th node are denoted as $f_i = \{x_i, d_i, e_i, l_i\}$, where $x_i$ are the coordinates, $d_i$ is the demand, and $e_i$ and $l_i$ are the early and late time windows, respectively. The input features are embedded through a linear projection to generate the initial feature embedding $h_i$. In each MHA layer, skip connections~\citep{he2016deep} and instance normalization (IN) are used.

\begin{equation}
    \begin{gathered}
        \hat{h}_i^{(l)}=IN^l\left(h_i^{(l-1)}+M H A^{(l)}\left(h_1^{(l-1)}, \ldots, h_n^{(l-1)}\right)\right), \\
        h_i^{(l)}=IN^l\left(\hat{h}_i+F F^l\left(\hat{h}_i\right)\right),
    \end{gathered}
\end{equation}
where $l$ and $l-1$ represent the current and last MHA layers, respectively. The FF contains a hidden sublayer with ReLU activations.
The above encoding process generates the final node embeddings $h_i^{(N)}$. This encoding is performed only once, and the static node embeddings are reused for every decoding step.

\textbf{Decoder} The decoder constructs a solution sequentially.  The input of the decoder includes three parts: the node embedding $h_1^{(N)}, \ldots, h_n^{(N)}$, the embedding of currently visited node $h_t^{(N)}$, and the attribute embedding $a_t$. All the node embeddings are produced by the decoder. $a_t$ is the embedding of the current state of attributes. We provide an attribute vector to include various attributes involved in multiple VRPs. At the $t$-th step, the attribute vector is $a_t = \{c_t,t_t,l_t,o_t\}$, where $c_t$ is the remaining capacity of the current vehicle, $t_t$ is the current time, $l_t$ is the current duration of the route, and $o_t$ indicates whether the route is open or not. Except for $o_t$, the others will be updated in each step. Backhauls are not embedded because they are implicitly considered in the node demands.

The decoder consists of one MHA layer and one single-head attention (SHA) layer with clipping. The MHA is slightly different from that used in the encoder. Skip connection, instance normalization, and FF sublayer are not used.

\begin{equation}
    \begin{gathered}
        \hat{h}_c = MHA_c \left( h_1^{(N)}, \ldots, h_n^{(N)}, h_t^{(N)}, a_t \right),\\
        u_1 \dots, u_n = SHA_c \left( h_1^{(N)}, \ldots, h_n^{(N)}, \hat{h}_c \right).\\
    \end{gathered}
\end{equation}

The output embedding of MHA $\hat{h}_c$ is used as the input of the SHA, and the SHA outputs the probabilities of choosing the next node using a softmax $p_i =\frac{e^{u_{ i}}}{\sum_j e^{u_{j}}}$. We omit the step indicator $t$ for readability. The detailed structure of MHA and SHA can be found in~\citet{kool2018attention}.

In each step, we need to mask some nodes from being selected. We update a masking vector $m_t$. The associated positions of unwanted nodes in the vector will be set to -inf, which will be used in the attentions before softmax. Except for masking those nodes that have already been selected in the previous steps, the infeasibility caused by various attributes should also be considered. For example, these nodes that violate time window constraints should not be selected.


\subsection{Attribute Composition} 

The input of the attribute composition is the input node features $f_i,i=1,\dots,n$, the list of visited nodes $V_t$ at the current step $t$, and the problem attributes $A$. The output is the attribute vector $a_t$ and the mask vector $m_t$.

The problem attributes $A$ are given explicitly with the input problem. $A$ are used to activate the corresponding attribute updating procedure in the attribute composition block. In this paper, we have four procedures for the four attributes. Each procedure $j$ updates the corresponding attribute in the attribute vector $a_t^j$ and calculates an infeasible node list that must not be visited in the next step $m_t^j$. The output attribute vector will include all the updated activated attributes and pad the inactivated attributes to be a default value. The output mask vector $m_t$ is the union of all activated attribute masks $m_t = V_t \cup \bigcup_{j \in A}m_t^j$. See Appendix~\ref{sec:a} for the details of the attribute procedures.

For example, if only the capacity attribute is involved in the problem (i.e., CVRP), the indicator only activates the capacity updating procedure. In each step, the remaining capacity of the current vehicle is calculated. The attribute vector $a_t = \{c_t,t_t,l_t,o_t\}$ only updates $c_t$, and pads the rest attributes to zero. The infeasible nodes that exceed vehicle capacity when added to the route are updated to update masking $m_t = V_t \cup m_t^c$.

Most of the investigated VRPs involve subsets of attributes in our unified model. We are learning the shared underlying attributes of diverse VRPs. In this way, we can train on a few VRPs and solve a much larger group of VRPs as arbitrary combinations of the underlying attributes. This characteristic enables the zero-shot generalization ability of our model.





\subsection{Multi-Task Reinforcement Learning}

We use the REINFORCE algorithm with a shared baseline following~\citet{kwon2020pomo}. We use greedy inference, i.e., a deterministic trajectory is constructed iteratively based on the policy. In each iteration, the next node is selected as the node with the maximum probability predicted by the decoder. $n$ trajectories are constructed from $n$ different starting points. Long-term rewards $R(\tau_1),\dots,R(\tau_n)$ (negative of the total distances) are calculated after the entire trajectories $\tau_1,\dots,\tau_n$ are constructed. For the model with parameters $\theta $, the following gradient ascent is used:

\begin{equation}
    \nabla_\theta J(\theta) \approx \frac{1}{nB} \sum_{i=1}^B\sum_{j=1}^n\left(R\left(\boldsymbol{\tau}^i_j \mid s_k^i \right)-b^i \left(s_k^i\right) \right) \nabla_\theta \log p_\theta\left(\boldsymbol{\tau}^i_j \mid s_k^i \right),
\end{equation}
where $s_k$ represents the instances are generated from $k$-th task (VRP). $p_\theta(\boldsymbol{\tau}^i_j)$ is the aggregation of the probability of selection in each step of the decoder. $b^i(s_k)=\frac{1}{n}\sum_{j=1}^n \left(R\left(\boldsymbol{\tau}^i_j\right)\right)$ is the shared baseline. $B$ is the batch size.

For multi-task learning, many optimizers have been designed to improve robustness and convergence. Instead of using sophisticated multi-task optimizers~\citep{chen2018gradnorm,kendall2018multi,zhang2021survey}, we simply trained the multiple tasks with equal weight.

\section{Experiments}

We conduct experiments on eleven vehicle routing problems, namely CVRP, VRPTW, OVRP, VRPB, VRPL, VRPBTW, VRPBL, OVRPL, OVRPLTW, OVRPBTW, and OVRPBLTW, along with benchmark datasets and our real-world logistic application cases. We train a unified model on the former five VRPs simultaneously and use the model to solve the rest problems in a zero-shot manner. Many of these routing problems are being addressed by the neural method for the first time. Note that one can easily extend the model to consider other attributes. We chose these attributes because they are among the most frequently used ones~\citep{braekers2016vehicle}.

\textbf{Instance Generation} We use the same problem setup as that used in~\citet{kool2018attention} to generate the basic CVRP. For VRPTW, we use the method introduced in~\citet{zhao2020hybrid} to generate time windows and service times. For the rest problems, there is no existing work that solves exactly the same settings. We make the following settings: For VRPB, we first generate a CVRP and then randomly select 20\% of customers as backhaul customers, whose demands are set to be the negative values of the original demands. For OVRP, we only need to set the open route indicator as active. For VRPL, the same maximum duration limit of $3$ is used for each route.

\textbf{Model Setting} 
The number of MHA for the encoder is 6, and the number of heads is 8. The hidden layer size is 512, and the embedding size is 128.

\textbf{Training} 
In training, we randomly select one type of VRP in each batch and generate instances of the selected VRP. We use 10,000 training instances for each epoch with a batch size of 64, and the number of epochs for training is 10,000. Adam optimizer is used. The initial learning rate is 1e-4 with a weight decay of 1e-6. We implement the unified model using PyTorch, and the experiments are running on a single RTX 2080Ti GPU. Training on the vehicle routing problems of size 100 costs about ten days.


\subsection{Results on Eleven VRPs}
We initially provide an overall performance comparison on all eleven VRPs and then introduce detailed settings and results on both the training and unseen VRPs. 

\begin{figure}[t]
\centering
    \begin{subfigure}{0.53\linewidth}
        \includegraphics[width=0.98\linewidth]{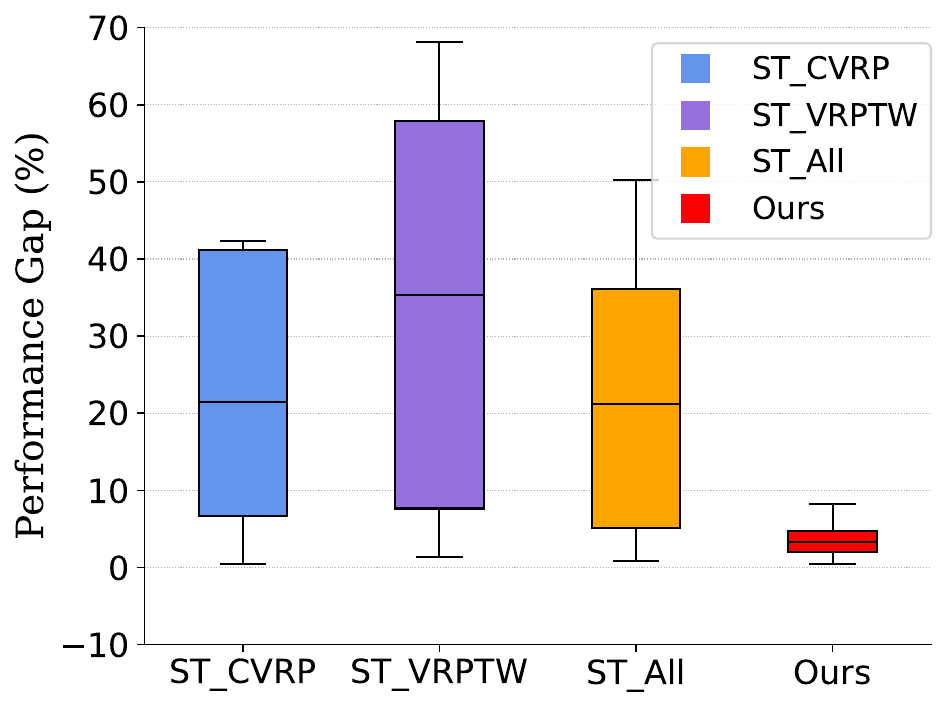}
    \end{subfigure}
    \begin{subfigure}{0.46\linewidth}
        \includegraphics[width=0.98\linewidth]{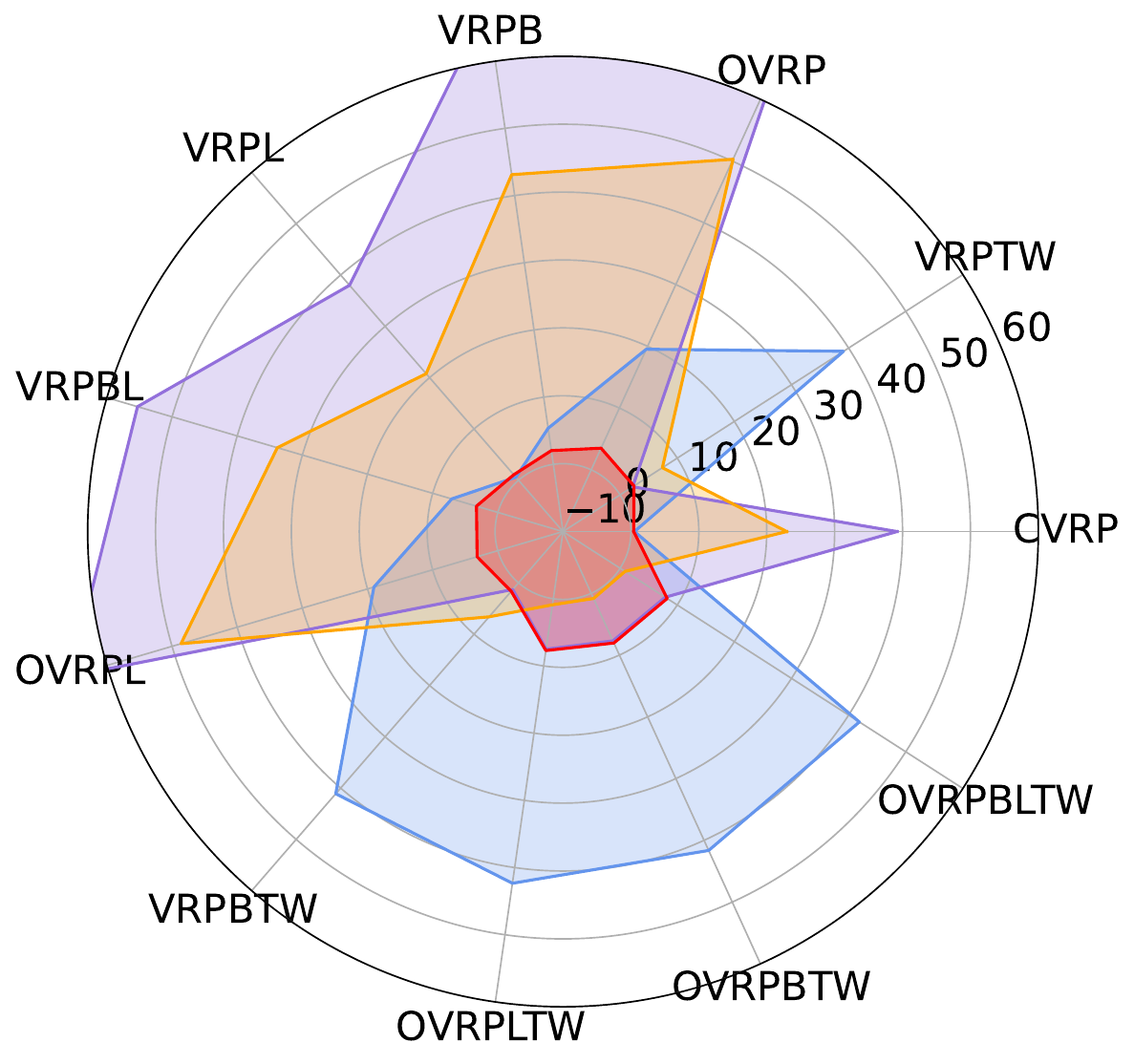}
    \end{subfigure}
    \caption{A comparison of gaps on eleven VRPs (Left: box plot, Right: radar plot). \textbf{ST} represents the unified model trained with single-task learning on CVRP, \textbf{ST\_all} represents the unified model with single-task learning on OVRPBLTW, and \textbf{MT} represents our approach, i.e., the unified model with multi-task learning on five VRPs. \textbf{ST\_FT} and \textbf{MT\_FT} are the fine-tuning models.}~\label{fig:gap}
\end{figure}

Figure~\ref{fig:gap} compares the performance in terms of the average gap (\%) to the baseline solver hybrid genetic search (HGS)~\citep{vidal2013hybrid}. 5,000 instances of size 50 are used for each problem. The box plot on the left displays the overall performance, while the radar plot on the right illustrates the average gap on each VRP, with a smaller area indicating better results. The red one represents the results of our unified model trained through multi-task learning with attribute composition. The three compared models in different colors have the same encoder-decoder structure and training settings as our unified model. The difference is that they are trained on a single problem. ST\_CVRP and ST\_VRPTW were trained on CVRP and VRPTW, respectively, while ST\_All was trained on OVRPBLTW with all attributes considered. 

Our unified model significantly outperforms single-task learning models with an average gap of less than 4\%. It demonstrates strong generalization capabilities across different VRPs. In comparison, single-task learning models only perform well on the training problem. For example, according to the results depicted in the radar plot, ST\_All achieves promising performance on the problem it was trained on (OVRPBLTW), as well as on two VRP variants with similar attributes (OVRPBTW and OVRPLTW), even slightly surpassing our unified model. However, its performance deteriorates significantly on the remaining VRPs. It suggests that training with all attributes fails to generalize effectively to other variants with only a subset of attributes. 


\subsubsection{Attribute Correlation}

We conduct a comparison of the distributions of different routing problems on a two-dimensional reduction space to visualize and analyze the attribute correlation. We first extract the sample features from the decoder hidden layer of our unified model. Each sample feature is the decoder hidden layer embedding of one inference step. For each VRP, we collect 1,000 feature samples. All the feature samples from different VRPs are transformed into one two-dimensional reduction space using t-distributed Stochastic Neighbor Embedding (t-SNE). 

\begin{figure}[t]
    \centering
    \includegraphics[width=0.9\linewidth]{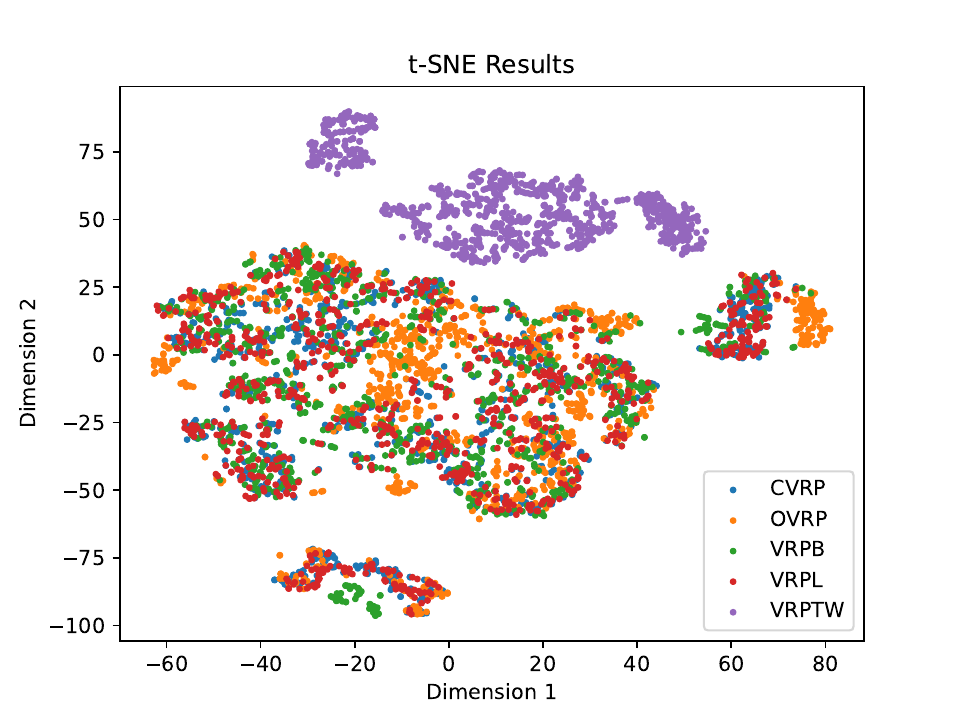}
    \caption{A comparison of distributions of different VRPs on two-dimensional reduction space of the decoder hidden layer.}~\label{fig:space}
\end{figure}

Figure~\ref{fig:space} shows a comparison of the distributions of five VRPs on the reduction space. There is a clear distinct distribution of VRPTW compared with others as the time window attribute poses a strong constraint over the route. In addition to VRPTW, OVRP also shows a different pattern than others because it does not force routes back to the depot presenting less constraint. In contrast, CVRP and VRPL follow a very close distribution and overlap with each other in most areas. This can be attributed to the fact that our route length limit is set to 3, which is easily satisfied in our settings. 

\begin{figure}[tbp]
    \centering
    \includegraphics[width = 1\linewidth]{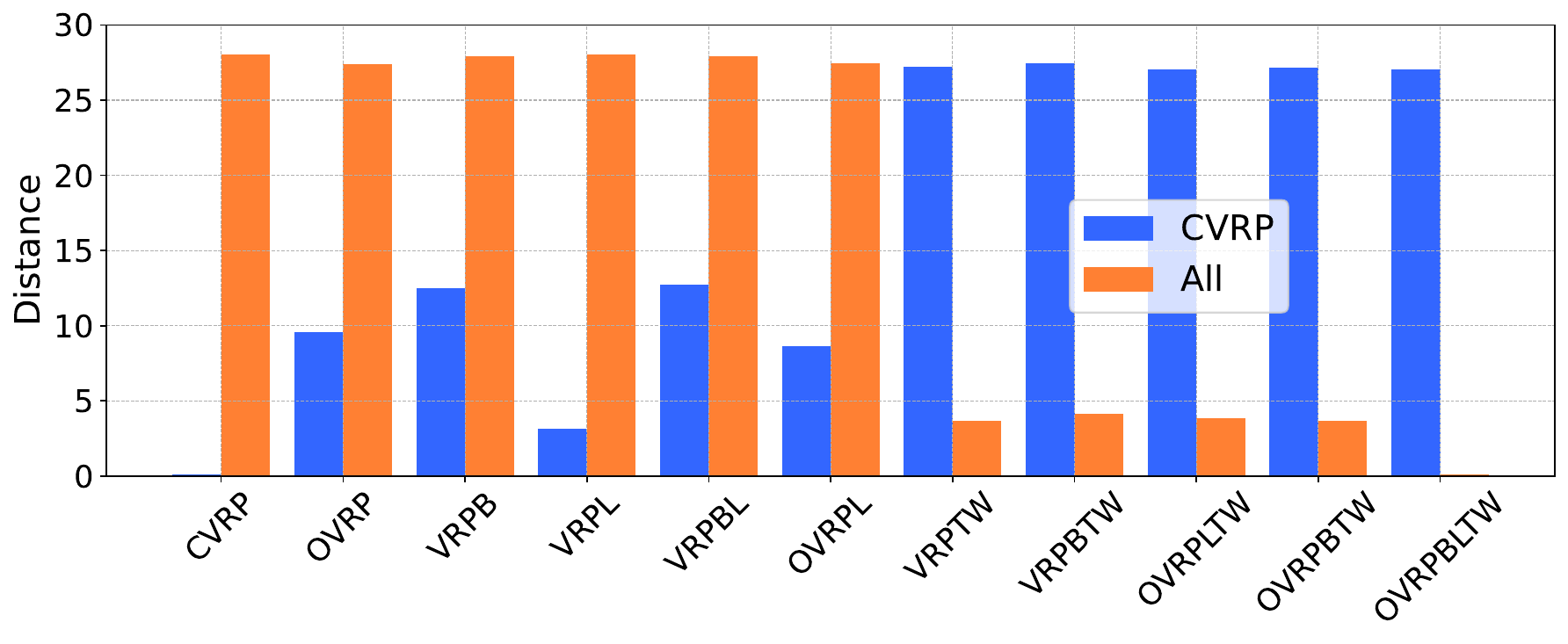}
    \caption{Hausdorff distance in reduction space comparing CVRP and OVRPBLTW (All) with other problems}~\label{fig:dis}
\end{figure}

We employ the Hausdorff distance to quantify the similarity between two VRPs in the reduced space. Figure~\ref{fig:dis} illustrates the Hausdorff distances of CVRP and OVRPBLTW (All) to other VRPs. The distances observed in the reduced space are consistent with the zero-shot generalization performance depicted in the radar plot in Figure~\ref{fig:gap}, with further details provided in Appendix~\ref{sec:b}. For instance, the model trained on CVRP exhibits a substantial distance from VRPs with time window constraints, correlating with its subpar performance on these problems. Furthermore, the optimal gap on VRPL coincides with the minimal distance between CVRP and VRPL.

\subsubsection{Training VRPs}


Table~\ref{table:performtrain} lists the experimental results on the five VRPs used in the training. For CVRP, we compare our results with the original version of AM~\citep{kool2018attention} and SOTA extensions, namely, POMO~\citep{kwon2020pomo}, and GCAM~\citep{zhu2023accelerated}. For VRPTW, a deep reinforcement learning method (DRL)~\citep{zhao2020hybrid} and POMO are compared. The results of LHK3~\citep{helsgaun2017extension} and Gurobi are used as the baseline for CVRP and VRPTW, respectively. For the rest three problems, there are no existing end-to-end NCO methods for comparison. We implement POMO on these problems and use the state-of-the-art hybrid genetic search (HGS)~\citep{vidal2013hybrid} solver as the baseline. 

\begin{table}[t]
\centering
\renewcommand{\arraystretch}{1.2}
\caption{Training costs between POMO with single-task learning and our model with multi-task learning on five VRPs
}~\label{table:trainingcost}
\resizebox{0.5\textwidth}{!}{%
\begin{tabular}{lccccc}
\toprule
     & \#Models & \#Total Para.  & \#Total Epochs & Time Cost (day) \\
     \midrule
POMO & 5        & 6.56M        & 50,000            & 49                       \\
Ours & 1        & 1.35M        & 10,000            & 10.5                     \\ 
\bottomrule
\end{tabular}
}
\end{table}

\begin{table}[htbp]
\centering
\renewcommand{\arraystretch}{1.05}
\caption{Experimental results on five training VRPs. (The compared neural solvers require training one model for each VRP)}\label{table:performtrain}
\resizebox{0.48\textwidth}{!}{%

\begin{tabular}{llcccccc}
\toprule
\multicolumn{1}{c}{\multirow{2}{*}{Problem}} & \multirow{2}{*}{Method} & \multicolumn{3}{c}{N=50} & \multicolumn{3}{c}{N=100} \\
\multicolumn{1}{c}{} & & Dis.   & Gap     & Time  & Dis.   & Gap& Time  \\
\midrule
CVRP     & HGS   & 10.38  & - & 7h    & 15.54  & -  & 14h   \\
   & LKH3  & 10.38  & 0.00\%  & 7h    & 15.61  & 0.46\%   & 14h   \\
   & AM (Samp1280)     & 10.59  & 2.02\%  & 7m    & 16.16  & 4.00\%   & 30m   \\
   & GCAM (Samp1280)   & 10.64  & 2.50\%  & -     & 16.29  & 4.83\%   & -     \\
   & POMO  & 10.53  & 1.41\%  & 3s    & 15.87  & 2.13\%   & 10s   \\
   & POMO (Aug8) & 10.44  & 0.58\%  & 15s   & 15.75  & 1.36\%   & 1.1m  \\
   & SGBS  & 10.39  & 0.12\%  & 2.0m  & 15.63  & 0.62\%   & 11.8m \\
   & Ours  & 10.56  & 1.73\%  & 3s    & 15.90  & 2.29\%   & 11s   \\
   & Ours (Aug8) & 10.47  & 0.85\%  & 20s   & 15.80  & 1.71\%   & 1.2m  \\
   & Ours+SGBS   & 10.40  & 0.18\%  & 2.3m  & 15.66  & 0.81\%   & 12.6m \\
   \midrule
VRPTW    & HGS   & 16.30  & - & 7h    & 26.14  & -  & 14h   \\
   & LKH3  & 16.52  & 1.36\%  & 7h    & 26.60  & 1.76\%   & 14h   \\
   & DRL (BS10)  & 17.90  & 9.82\%  & 1m    & 29.50  & 12.85\%  & 2m    \\
   & DRL (BS10) +LNS   & 16.94  & 3.93\%  & 11m   & 27.44  & 4.97\%   & 65m   \\
   & POMO  & 16.78  & 2.97\%  & 3s    & 27.13  & 3.77\%   & 11s   \\
   & POMO (Aug8) & 16.66  & 2.22\%  & 19s   & 26.91  & 2.93\%   & 1.2m  \\
   & SGBS  & 16.55  & 1.52\%  & 2.9m  & 26.55  & 1.58\%   & 15.1m \\
   & Ours  & 16.96  & 4.06\%  & 3s    & 27.46  & 5.05\%   & 11s   \\
   & Ours (Aug8) & 16.80  & 3.09\%  & 20s   & 27.13  & 3.81\%   & 1.2m  \\
   & Ours+SGBS   & 16.58  & 1.71\%  & 3.2m  & 26.63  & 1.89\%   & 17.9m \\
   \midrule
OVRP     & HGS   & 6.49   & - & 7h    & 9.71   & -  & 14h   \\
   & LKH3  & 6.52   & 0.46\%  & 7h    & 9.75   & 0.41\%   & 14h   \\
   & POMO  & 6.73   & 3.67\%  & 3s    & 10.18  & 4.91\%   & 10s   \\
   & POMO (Aug8) & 6.63   & 2.14\%  & 16s   & 10.07  & 3.76\%   & 1.1m  \\
   & SGBS  & 6.56   & 1.12\%  & 2.1 m & 9.89   & 1.92\%   & 12.1m \\
   & Ours  & 6.81   & 4.90\%  & 3s    & 10.34  & 6.56\%   & 11s   \\
   & Ours (Aug8) & 6.71   & 3.40\%  & 20s   & 10.14  & 4.48\%   & 1.2m  \\
   & Ours+SGBS   & 6.59   & 1.58\%  & 2.5m  & 9.94   & 2.38\%   & 13.4m \\
   \midrule
VRPB     & HGS   & 7.69   & - & 7h    & 11.13  & -  & 14h   \\
   & LKH3  & 7.70   & 0.18\%  & 7h    & 11.29  & 1.40\%   & 14h   \\
   & POMO  & 7.92   & 3.06\%  & 3s    & 11.57  & 3.88\%   & 10s   \\
   & POMO (Aug8) & 7.84   & 2.05\%  & 15s   & 11.43  & 2.68\%   & 1.1m  \\
   & SGBS  & 7.78   & 1.22\%  & 1.9m  & 11.31  & 1.59\%   & 11m   \\
   & Ours  & 8.17   & 6.36\%  & 3s    & 11.72  & 5.23\%   & 11s   \\
   & Ours (Aug8) & 7.87   & 2.40\%  & 20s   & 11.53  & 3.58\%   & 1.2m  \\
   & Ours+SGBS   & 7.78   & 1.25\%  & 2.1m  & 11.36  & 2.06\%   & 11.7m \\
   \midrule
VRPL     & HGS   & 10.37  & - & 7h    & 15.54  & -  & 14h   \\
   & LKH3  & 10.37  & 0.03\%  & 7h    & 15.61  & 0.43\%   & 14h   \\
   & POMO  & 10.55  & 1.78\%  & 3s    & 15.84  & 1.96\%   & 10s   \\
   & POMO (Aug8) & 10.46  & 0.91\%  & 16s   & 15.72  & 1.14\%   & 1.1m  \\
   & SGBS  & 10.40  & 0.30\%  & 2.3m  & 15.64  & 0.66\%   & 13.1m \\
   & Ours  & 10.56  & 1.88\%  & 3s    & 15.96  & 2.72\%   & 11s   \\
   & Ours (Aug8) & 10.47  & 0.98\%  & 20s   & 15.80  & 1.66\%   & 1.2m  \\
   & Ours+SGBS   & 10.40  & 0.33\%  & 2.6m  & 15.67  & 0.83\%   & 14.3m \\
   \midrule
Average  & POMO  & 10.50  & 2.58\%  & 3s    & 16.12  & 3.33\%   & 10s   \\
   & POMO (Aug8) & 10.41  & 1.58\%  & 16s   & 15.97  & 2.37\%   & 1.1m  \\
   & SGBS  & 10.34  & 0.86\%  & 2.24m & 15.81  & 1.28\%   & 12.6m \\
   & Ours  & 10.61  & 3.78\%  & 3s    & 16.27  & 4.37\%   & 11s   \\
   & Ours (Aug8) & 10.46  & 2.14\%  & 20s   & 15.81  & 3.05\%   & 1.2m  \\
   & Ours+SGBS   & 10.35  & 1.01\%  & 2.5m  & 15.85  & 1.59\%   & 14.0m \\
   \bottomrule
\end{tabular}%

}
\end{table}

\begin{table*}[t]
\centering
\small
\renewcommand{\arraystretch}{1.1}
\caption{Zero-shot generalization performance on five new VRPs.}
\label{table:performunseen}
\resizebox{0.8\textwidth}{!}{%
\begin{tabular}{clcccc|clcccc}

\toprule
\multirow{2}{*}{VRP} & \multirow{2}{*}{Method} & \multicolumn{2}{c}{n50} & \multicolumn{2}{c|}{n100} & \multirow{2}{*}{VRP} & \multirow{2}{*}{Method} & \multicolumn{2}{c}{n50} & \multicolumn{2}{c}{n100} \\
\multicolumn{1}{c}{}                         & \multicolumn{1}{c}{}                        & Dis.      & Gap         & Dis.       & Gap         & \multicolumn{1}{c}{}                         & \multicolumn{1}{c}{}                        & Dis.      & Gap         & Dis.       & Gap         \\
\midrule

\multirow{6}{*}{\rotatebox{90}{VRPBL}} & HGS & 7.70      & -           & 11.15      & -           &  \multirow{6}{*}{\rotatebox{90}{OVRPLTW }}    & HGS & 10.69     & -           & 17.35      & -           \\
 & NI  & 11.69     & 51.86\%     & 17.38      & 55.92\%     &  & NI  & 15.74     & 47.20\%     & 26.16      & 50.78\%     \\
 & FI  & 11.61     & 50.81\%     & 16.37      & 46.88\%     &  & FI  & 15.22     & 42.41\%     & 25.81      & 48.79\%     \\
 & POMO\_CVRP   & 8.21      & 6.61\%      & 12.41      & 11.37\%     &  & POMO\_CVRP   & 15.23     & 42.46\%     & 26.75      & 54.18\%     \\
 & POMO\_VRPTW  & 13.43     & 74.45\%     & 17.86      & 60.27\%     &  & POMO\_VRPTW  & 11.51     & 7.70\%      & 19.41      & 11.88\%     \\
 & Ours& \textbf{7.97 }     & \textbf{3.48\%  }    & \textbf{11.65  }    & \textbf{4.50\%  }    &  & Ours& \textbf{11.50 }    &\textbf{ 7.59\%  }    & \textbf{19.34 }     & \textbf{11.50\% }    \\
 \midrule
\multirow{6}{*}{\rotatebox{90}{OVRPL}}& HGS & 6.49      & -           & 9.71       & -           & \multirow{6}{*}{\rotatebox{90}{OVRPBTW}}     & HGS & 10.67     & -           & 17.31      & -           \\
 & NI  & 13.85     & 113.55\%    & 13.61      & 40.17\%     &  & NI  & 15.76     & 47.77\%     & 26.24      & 51.60\%     \\
 & FI  & 14.34     & 121.11\%    & 13.46      & 38.62\%     &  & FI  & 15.22     & 42.69\%     & 25.79      & 48.99\%     \\
 & POMO\_CVRP   & 7.75      & 19.47\%     & 11.78      & 21.35\%     &  & POMO\_CVRP   & 15.25     & 42.98\%     & 26.78      & 54.69\%     \\
 & POMO\_VRPTW  & 10.80     & 66.54\%     & 18.62      & 91.78\%     &  & POMO\_VRPTW  & 11.52     & 7.94\%      & 19.48      & 12.51\%     \\
 & Ours& \textbf{6.69 }     & \textbf{3.10\% }     &\textbf{ 10.15 }     & \textbf{4.57\% }     &  & Ours& \textbf{11.49}     & \textbf{7.72\%  }    & \textbf{19.32}      &\textbf{ 11.61\% }    \\
 \midrule
\multirow{6}{*}{\rotatebox{90}{VRPBTW}}      & HGS & 16.43     & -           & 26.31      & -           & \multirow{6}{*}{\rotatebox{90}{Average}}     & HGS & 10.39     & -           & 16.36      & -           \\
 & NI  & 18.92     & 15.15\%     & 36.84      & 40.05\%     &  & NI  & 15.19     & 46.16\%     & 24.05      & 46.95\%     \\
 & FI  & 18.33     & 11.56\%     & 36.59      & 39.10\%     &  & FI  & 14.95     & 43.78\%     & 23.60      & 44.25\%     \\
 & POMO\_CVRP   & 22.98     & 39.85\%     & 38.99      & 48.20\%     &  & POMO\_CVRP   & 13.88     & 33.56\%     & 23.34      & 42.64\%     \\
 & POMO\_VRPTW  & \textbf{16.63 }    & \textbf{1.22\% }     & 27.18      & 3.33\%      &  & POMO\_VRPTW  & 12.78     & 22.93\%     & 20.51      & 25.34\%     \\
 & Ours& 16.70     & 1.66\%     &\textbf{ 27.11}      & \textbf{3.05\% }     &  & Ours& \textbf{10.87}     & \textbf{4.59\%}      & \textbf{17.51}      & \textbf{7.03\% }   \\
 \bottomrule
\end{tabular}%

}
\end{table*}

We extended the original POMO on these problems and trained these models independently on each problem with single-task learning. We keep the settings of the original paper~\citep{kwon2020pomo}. For other compared single-task learning models, we select the best results from the corresponding papers. If additional inference techniques are used, such as beam search (BS) and sampling (Samp), their size is indicated in the parentheses following the method. We adopt additional data augmentation (Aug) following POMO~\citep{kwon2020pomo}. In addition, the advanced inference strategy simulation guided beam search (SGBS) has been investigated and integrated into our framework for better zero-shot generation performance. 

Tabel~\ref{table:trainingcost} presents the required number of models and the total parameter sizes for single-task learning and our multi-task learning for five tasks. The required number of models (and the total parameters) could be much higher for single-task learning if we train a different model for each possible attribute combination, which is not affordable for real-world applications.

For each problem, the evaluations are conducted on 5,000 instances. We compare the performance with respect to three criteria: the average distance (Dis.), the gap of the average distance to the baseline results, and the total running time on 5,000 instances. In general, our unified model is competitive with the existing single-task NCO methods. On CVRP and VRPTW, the results are better than the various existing end-to-end methods except for POMO. The gap between our model and the optimal baselines is less than 5\% across all five VRPs tested, and our running time is significantly less than the baseline methods. The average gap of our unified model on the five training VRPs can be further reduced to around 1\% when integrating with SGBS, with an acceptable increase in inference time. Despite slightly inferior results compared to POMO and SGBS, we note that the latter requires training individual neural networks for each problem and does not generalize well on unseen problems.

\subsubsection{Unseen VRPs}

We use our unified model to solve unseen VRPs in a zero-shot manner. The experiments are carried out on five VRPs (i.e., VRPBTW, VRPBL, OVRPL, OVRPLTW, and OVRPBTW).

We compare the results of our unified model with single-task models, two commonly used constructive heuristics, and the SOTA heuristic HGS. The constructive heuristics are the nearest insertion method and farthest insertion method. We extended the source code of HGS so that it can solve these routing problems~\citep{vidal2013hybrid}. The two single-task models are POMO trained on CVRP and VRPTW, respectively. We added the masking procedure used in our unified model to POMO so that they are applicable to different VRPs. Our unified model and the two single-task models are used in a zero-shot way without any fine-tuning on the new VRPs.

Table~\ref{table:performunseen} shows the zero-shot performance. The results are evaluated on 5,000 instances for each problem. Our unified model outperforms other methods including two heuristic methods except for HGS, which is specifically developed for VRPs. The two single-task models are inferior to our multi-task unified model. The deficiency of single-task models is more obvious in problems with very different attributes. For example, the model only trained on CVRP performs worse on these problems involving time windows attributes, while the model trained on VRPTW has poor performance on VRPBL. The average gap of our model over these five unseen VRPs is 4.6\% and 7\% on VRPs of sizes 50 and 100, respectively.

\subsection{Benchmark Datasets}

We apply our model on the well-known CVRPLib benchmark datasets to demonstrate the out-of-distribution performance. Most of them are derived from real-world instances and have diverse distributions and sizes. Specifically, we select six test suites with diverse attributes from CVRPLIB~\footnote{http://vrp.atd-lab.inf.puc-rio.br/}. There are a total of 181 instances whose problem size ranges from 30 to 1,000. We normalize the coordinates of customers so that they are within the unit range of $[0,1]$. We also normalize the demands with respect to the vehicle capacity.

To evaluate the performance, we compare our unified model with POMO, which is trained on CVRP. Both models are trained on instances of size 100. We use the best-known solutions (BKS) provided by CVRPLIB as the baseline. The criteria for comparison include the average distance over all test suite instances and the average gap to BKS.

Table~\ref{table:cvrplib} lists the information of six test suites as well as the experimental results. The results demonstrate that, although our model is not specifically trained for CVRP, it outperforms POMO on all test suites. Overall, the average gap between our unified model and the BKS is less than 10\%, which is about half of POMO. 

We divide the 100 instances in X into four groups based on their problem size. The results of our model and POMO on each group are summarized in Table~\ref{table:cvrplib}. Our findings indicate that both models perform similarly well on small-scale instances, but as the scale of the problem increases, the performance of POMO deteriorates. In fact, in the largest group, the average gap of POMO is over 30\%. Conversely, our model is more robust across problem sizes, with its gap increasing only from 6.2\% to 13.8\%.

\begin{table}[tbp]

\caption{Results on CVRPLib datasets with diverse distributions and sizes.}~\label{table:cvrplib}
\centering
\resizebox{0.5\textwidth}{!}{%
\begin{tabular}{lllllll}
\toprule
\multirow{2}{*}{Benchmark} & \multirow{2}{*}{Size} & \multicolumn{1}{c}{\multirow{2}{*}{BKS}} & \multicolumn{2}{c}{POMO} & \multicolumn{2}{c}{Ours}   \\
    & & \multicolumn{1}{c}{}  & Dis. & Gap & Dis.     & Gap  \\
    \midrule
Set A   & 31-79& 1041.9  & 1104.8 & 6.0\%     & 1066.8   & \textbf{2.4\%}  \\
Set B   & 30-77& 963.7   & 1065.7 & 10.6\%    & 992.2    & \textbf{2.9\%}  \\
Set F   & 44-134     & 707.7   & 770.6& 8.9\%     & 760.6    & \textbf{7.5\%}  \\
Set M   & 100-199    & 1083.6  & 1145.2 & 5.7\%     & 1141.9   & \textbf{5.4\%}  \\
Set P   & 15-100     & 587.4   & 660.4& 12.4\%    & 627.1    & \textbf{6.8\%}  \\
\midrule
\multirow{4}{*}{Set X}   & 100-300    & 33868.5 & 36299.07     & 7.2\%     & 35954.7  & \textbf{6.2\%}  \\
   & 300-500    & 63774.8 & 71524.15     & 12.2\%    & 68841.6  & \textbf{7.9\%}  \\
   & 500-700    & 88561.4 & 107927.9     & 21.9\%    & 98843.8  & \textbf{11.6\%} \\
   & 700-1000   & 113619.5& 149858.6     & 31.9\%    & 129270.9 & \textbf{13.8\%} \\
\bottomrule
\end{tabular}%
}
\end{table}

\subsection{Industry Logistic Application}


Finally, we validate the unified model on a real-world industry logistic application with three different cases. The test sets are derived from our historical cross-city transportation logistic data spanning 30 days. Three types of cases are examined. The first case involves the typical CVRP, where only the vehicle capacity constraint is taken into account. Three datasets containing instances of different vehicle capacities: large(L), middle(M), and small(S) were collected. These datasets consisted of a total of 64 instances, with customer numbers varying from 50 to 100 and demands ranging from 25 to the maximum vehicle capacity. In the second case OVRP, drivers are not required to return vehicles to the depot, resulting in an open route scenario. In the final case VRPL, the duration limit indicated by the route length is imposed on the drivers. In the last VRPL case, only one dataset is collected with the largest vehicle capacity, as duration limit constraints are typically not impactful on vehicles with small capacity. Table~\ref{table:industry} lists the statistics of the industry logistic datasets.

\begin{figure}
    \centering
    \includegraphics[width=0.48\textwidth]{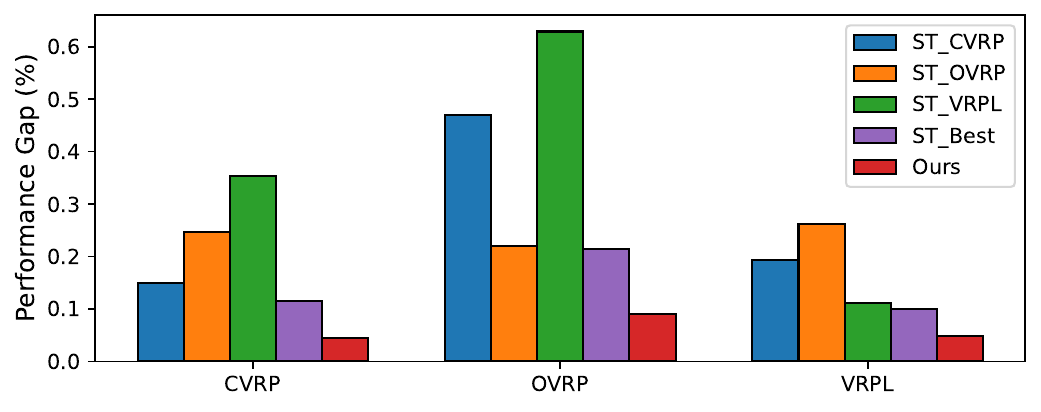}
    \caption{Results on real-world logistic application. CVRP, OVRP, and VRPL represent the three investigated application cases. ST\_CVRP, ST\_OVRP, and ST\_VRPL represent the model trained on CVRP, OVRP, and VRPL, respectively. ST\_Best denotes the best result among the three single-task learning models for each instance.}~\label{fig:industryresult}
    \label{fig:enter-label}
\end{figure}

We compared our unified model to models trained using single-task learning in a zero-shot manner, without any fine-tuning or adaptation. The three models under comparison were trained on CVRP, OVRP, and VRPL datasets, respectively. Figure~\ref{fig:industryresult} depicts the average gaps to the baseline HGS in these three cases. For CVRP and OVRP, the results are averaged across all three sets. ST\_Best denotes the average gaps of the best result among the three single-task learning models for each instance. Our unified model outperformed the best results of the compared models, achieving gaps of less than 5\% on CVRP and VRPL, and 10\% on OVRP. 

\begin{table}[tbp]
\centering
\caption{Real-world logistic application instance statistics}~\label{table:industry}
\small
\resizebox{0.32\textwidth}{!}{%
\begin{tabular}{ccccccc}
\toprule
                \multicolumn{2}{c}{Case}     & Size   & Capacity & Open & Limit \\
                      \midrule
\multirow{3}{*}{CVRP} & SetA            & 50-100 & L      & \ding{55}          & \ding{55}              \\
                      & SetB           & 50-100 & M     & \ding{55}          & \ding{55}              \\
                      & SetC            & 50-100 & S      & \ding{55}          & \ding{55}             \\
                      
\multirow{3}{*}{OVRP} & SetA            & 50-100 & L      & \textcolor{red}{\ding{51}}          & \ding{55}              \\
                      & SetB           & 50-100 & M      & \textcolor{red}{\ding{51}}          & \ding{55}              \\
                      & SetC           & 50-100 & S      & \textcolor{red}{\ding{51}}          & \ding{55}              \\
\multicolumn{2}{c}{VRPL}               & 50-100 & L      & \ding{55}          & \textcolor{red}{\ding{51}}             \\
\bottomrule
\end{tabular}%
}
\end{table}

\section{Conclusion}

This work has proposed a novel learning-based approach to solve various vehicle routing problems (VRPs) with different attributes by a single unified model. In our proposed framework, different VRPs are treated as different combinations of several shared basic attributes (e.g., time windows, open routes, backhauls, and duration limits). In this way, by leveraging the similarities and relations among different VRPs with shared underlying attributes, we can build a single unified model to tackle numerous VRPs in an end-to-end manner. To the best of our knowledge, this work is the first attempt to develop a cross-problem neural VRP solver, which is desirable for many real-world industrial applications. Experimental results demonstrate that our proposed model exhibits promising zero-shot generalization performance on unseen VRPs with new combinations of the basic attributes. Remarkably, without any fine-tuning, our model can outperform single-task models substantially, reducing the average performance gap from over 20\% to approximately 5\% across eleven VRPs, and achieving significant performance boost on benchmark datasets as well as a real-world logistics application.

\bibliographystyle{ACM-Reference-Format}
\bibliography{main}

\appendix

\section{Model Details}~\label{sec:a}

\subsection{Attention}
We use the attention mechanism in~\citet{vaswani2017attention}, which is a mapping (message passing~\citep{kool2018attention}) of query $Q$, key $K$, and value $V$ vectors to an output. For each node $i$, the query $Q_i$, key $K_i$, and value $V_i$ are projections of the input embedding $h_i$:
\begin{equation}
    Q_i = W^Qh_i, K_i = W^K h_i, V_i = W^V h_i.
\end{equation}
where the parameters $W^Q$ and $W^K$ are of size ($d_k \times d_h$) and $W^V$ is of size ($d_v \times d_h$). We compute the compatibility $u_{ij}$ from the queries and keys as follows:
\begin{equation}
    u_{ij} = \frac{Q_i^T K_j}{\sqrt{d_k}}.
\end{equation}
Then we scale the compatibilities $u_{ij}$ using softmax to get attention weights $a_{ij} \in [0,1]$:
\begin{equation}
    a_{ij} = \frac{e^{u_{ij}}}{\sum_j e^{u_{ij}}}.
\end{equation}
The output vector $h_i^o$ for node $i$ is the combination of the weights $a_{ij}$ and values $V_j$:
\begin{equation}
    h_i^o = \sum_j a_{ij} V_j.
\end{equation}

\subsection{Multi-Head Attention (MHA)}

Multi-head attention enables the model to learn diverse information and usually benefits the results. MHA consists of $h$ heads and each head is an attention. It concatenates the results from all heads with a linear projection.
\begin{equation}
    \begin{aligned}
        MHA (h_1,\dots,h_n) &= Concat (head_1,\dots,head_h) W^O \\
        head_i &= Attention (h_1,\dots,h_n)
    \end{aligned}
\end{equation}
where $W^O$ has size ($hd_v \times d_k$). In our experiments, we use 8 heads with different parameters, and the embedding size is 128. For the attention model in each head, the parameter dimensions are $d_k=d_v=d_h/h=16$.

\subsection{Decoder Details}

We use an MHA followed by a SHA in the decoder following~\citet{kool2018attention}. The computing of queries, keys, and values for the MHA are as follows:
\begin{equation}
    \begin{aligned}
     Q_c &= W^Qh_c, K_i = W^K h_i, V_i = W^V h_i, \\
        h_c &= Concat(h_{t},a_t),      
    \end{aligned}
\end{equation}
where $h_{t}$ is the embedding of the current visited node and $a_t$ is the attribute vector. $h_i$ is the output embedding from the encoder for node $i$.

In the SHA, we compute the compatibility using equation (5) and clip the results within [-10,10] with tanh. We also exclude the masked nodes by setting their compatibility values to -inf:
\begin{equation}
    u_{cj} = \left\{
    \begin{aligned}
     & 10 \cdot tanh \left( \frac{q_{c}^T k_j}{\sqrt{d_k}}\right) \quad &\text{if } j \notin m_t \\ 
     & -inf \quad &\text{otherwise}    
    \end{aligned}
    \right.
\end{equation}
The output probability of selecting next node is computed as the softmax of the output compatibilities $p_i =\frac{e^{u_{ i}}}{\sum_j e^{u_{j}}}$.

\subsection{Attribute Procedures}

\textbf{Capacity}
We track the remaining vehicle capacity $c_t$ at each step $t$, which is initially set to be the capacity of the vehicle $c_1 = 1$ (all demands have been scaled by the capacity). After selecting a new node $v_t$, we update the remaining capacity as:
\begin{equation}
    c_t = c_{t-1} - d_t
\end{equation}
where $c_{t-1}$ represents the remaining capacity from the previous step, and $d_t$ represents the demand of the selected node in the current step $t$.

We mask these nodes that have already been visited or have demands that exceed the remaining vehicle capacity.

\textbf{Time windows}
We keep track of the current time $t_t$ at each step $t$, initialized as $t_1 = 0$. After the selection of a new node, we update the current time as:
\begin{equation}
    t_t = max (t_{t-1} + c_{(t-1),t} , e_t) + s_t
\end{equation}
where $t_{t-1}$ represents the time from the previous step. $c_{(t-1),t}$ represents the distance between the last node and the current node (i.e., the traveling time cost between two nodes). $e_t$ and $s_t$ represent the early time window and the service time at node $v_t$, respectively.

We mask the visited nodes and the nodes whose time windows cannot be satisfied: 1) when we are unable to visit the node within the feasible time windows starting from the current node, or 2) when visiting the node would result in being too late to return back to the depot.

\textbf{Duration limit}
We keep track of the route length $l_t$ at each step $t$, which is initialized to be zero $l_t = 0$. We update the current route length by adding the route length from the previous step $t-1$ and the distance between the last node and the current node:
\begin{equation}
    l_t = l_{t-1} + c_{(t-1),t}
\end{equation}
We mask the nodes that exceed the duration limit when selected.

\textbf{Open route}
We only need a fixed binary indicator for the open route attribute, with $o_t=1$ representing that the route is open and $o_t=0$ otherwise. It does not contribute to the masking vector. 

However, unlike other attributes, the open route attribute results in a different total distance calculation. The distance between the last node and the depot is not included. 

\section{Results on Eleven VRPs}~\label{sec:b}

We present detailed results on eleven VRPs to demonstrate the advantages of our unified model with multi-task learning and attribute composition. We compared the following settings:
\begin{itemize}
    \item Our unified model with multi-task learning.
    \item Our unified model with single-task learning on each training VRP.
    \item Our unified model with single-task learning on the VRP with all training attributes.
    \item POMO with single-task learning on each training VRP
\end{itemize}

We also examined the influence of the normalization method and included the results with fine-tuning. Table~\ref{table:ablation} lists all the results for VRPs of size 50. The top three results for each VRP are highlighted \textbf{in bold}. The abbreviations \textbf{ST} and \textbf{MT} represent single-task learning and our multi-task learning, respectively. \textbf{NN},\textbf{ BN}, \textbf{IN}, and \textbf{RN} denote no normalization, batch normalization, instance normalization, and re-zero normalization, respectively. \textbf{FT} represents fast fine-tuning. All models were trained on a single RTX 2080Ti GPU. The testing time cost is the duration it takes to solve 5,000 instances for each VRP. 

\textbf{MT vs. ST} 
It is clear from this table that the single-task learning baselines perform similarly to the POMO counterparts. These results come without surprise since the POMO model serves as our base model. The performance of our multi-task learning model is slightly poorer than POMO and single-task learning (which is directly trained on each specific task), but our model has a much better (zero-shot) generalization performance on other seen/unseen tasks. Therefore, our method has a much better average performance across multiple VRP variants. In addition, we only need to build and train one single model to tackle all tasks with various attribute combinations, while single-task learning needs to build a different model for each task. The training budget has been significantly reduced.

\textbf{Attribute Composition} As discussed in the main paper, we directly train the unified model on OVRPBLTW with all attributes taken into account (ST\_OVRPBLTW) to demonstrate the benefit of attribute composition. Line 11 in Table~\ref{table:ablation} lists the results of ST\_OVRPBTW. According to the results, ST\_OVRPBLTW can achieve promising performance on the problem it trained on (OVRPBLTW) as well as two VRP variants with similar attributes (OVRPBTW and OVRPLTW). However, its performance becomes extremely poor for the rest VRPs (from simple CVRP to VRPBTW). This observation suggests that ST\_OVRPBLTW is over-fitted to CVRPBLTW with all attributes and cannot generalize well to other variants with a subset of attributes. This comparison also confirms the effectiveness and usefulness of our proposed method for learning different VRP variants with attribute composition.

\textbf{Normalization} Line 13 to 16 list the outcomes of the multi-task learning experiment without normalization, batch normalization, instance normalization, and re-zero normalization (BQ-NCO). Overall, we observed that normalization techniques had minimal impact on the results. Surprisingly, even without normalization, the results were already satisfactory. However, it is important to mention that these tests were conducted solely on our unified model with VRPs of size 50. In future work, a comprehensive study is required.

\textbf{Fine-tuning}  We conducted fast fine-tuning of ST\_CVRP and MT on eleven VRPs. The fine-tuning on each problem costs less than two hours on size 50 and about 5 hours on size 100. The results indicate that the unified model with fine-tuning can yield further performance improvement. With fine-tuning, our unified model achieves the best results with a small gap across all VRPs. See Appendix \ref{sec:c} for the experimental settings of fine-tuning and more results.

\section{Fast Fine-Tuning on Unseen VRPs}~\label{sec:c}

We conduct experiments to show the performance of our model under fast fine-tuning. The same five unseen VRPs are used: VRP with backhauls and time windows (VRPBTW), VRP with backhauls and duration limitation (VRPBL), open VRP with duration limitations (OVRPL), open VRP with duration limitations and time windows (OVRPLTW), and open VRP with backhauls and time windows (OVRPBTW).

Two different fine-tuning settings are tested: 1) only updating the decoder while keeping the encoder fixed, and 2) updating the entire model. Each epoch is trained using 10,000 instances with a batch size of 64, and 200 epochs are used. The learning rate and weight decay are set to 1e-5 and 1e-6, respectively. The experiments are carried out on the instances of size 100. The entire fine-tuning process with 200 epochs takes approximately five hours, while we note that the model typically converges within the first 50 epochs.

\begin{figure}[t]
\centering
     \includegraphics[width=0.9\linewidth]{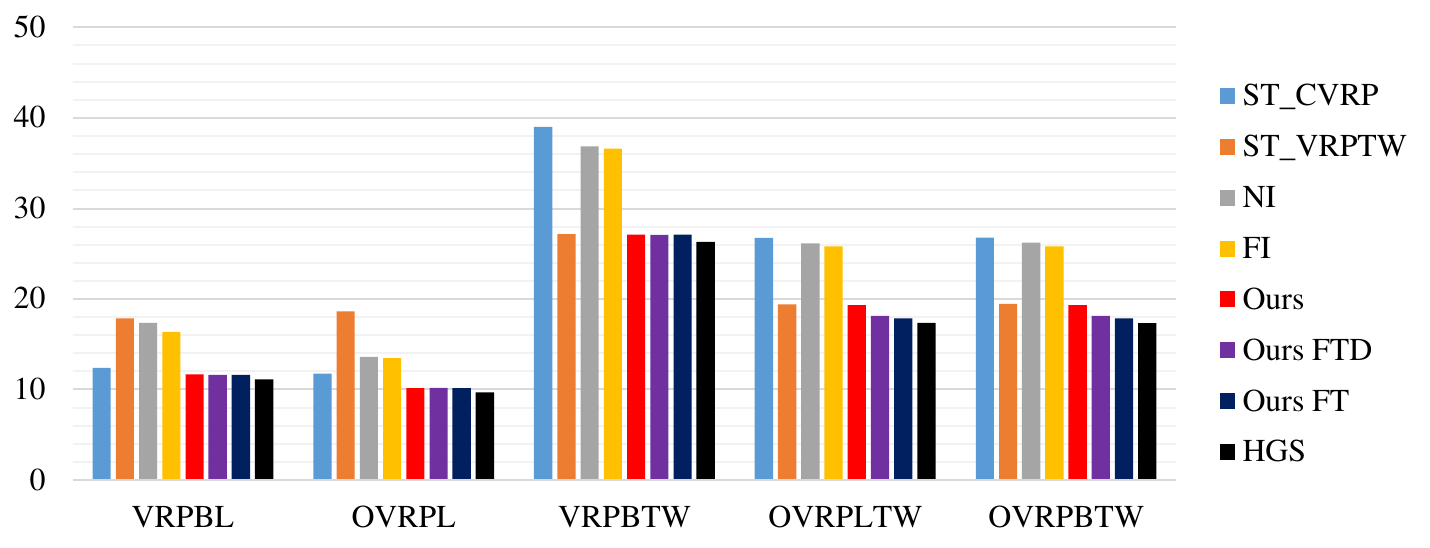}
     \caption{Comparison of the average results (total distances) of different methods on five new VRPs. }\label{fig:newvrps}
\end{figure}

Figure~\ref{fig:newvrps} provides a comparison of different methods on the five VRPs in terms of distance. The detailed results are listed in Table~\ref{table:newvrps}, where the distance (Dis.), gap to the baseline HGS, and running time of methods are compared on 5,000 instances. The best results are shown in bold. 

The abbreviations POMO\_CVRP and POMO\_VRPTW denote the single-task model (POMO) trained on CVRP and VRPTW instances, respectively. NI and FI represent the nearest and farthest insertion heuristics, and HGS is the SOTA algorithm for VRPs. FTD and FT indicate fine-tuning on the decoder and the entire network, respectively.

\newpage

\begin{landscape}

\begin{table}[htbp]
\renewcommand{\arraystretch}{1.8}
\centering
\Large
\caption{A summary of results on eleven VRPs with size 50. The top three results on each VRP are \textbf{in bold}. \textbf{ST} and \textbf{MT} represent single-task learning and the proposed multi-task learning, respectively. \textbf{NN},\textbf{ BN}, \textbf{IN}, and \textbf{RN} represent no normalization, batch normalization, instance normalization, and re-zero normalization, respectively. \textbf{FT} represents fast fine-tuning. All the models are trained on a single RTX 2080Ti GPU. The testing time cost is the time cost of solving 5,000 instances on each VRP. }~\label{table:ablation}

\resizebox{\linewidth}{!}{%
\begin{tabular}{cclcccccccccccccc}
\hline
\multicolumn{3}{c}{\multirow{2}{*}{Method}}          & \multicolumn{2}{c}{Time Cost}         & \multicolumn{12}{c}{Gap}          \\ \cline{4-17} 
\multicolumn{3}{c}{}   & Training      & Testing               & CVRP            & VRPTW           & OVRP            & VRPB            & VRPL            & VRPBL           & OVRPL           & VRPBTW          & OVRPLTW         & OVRPBTW         & OVRPBLTW        & Average         \\ \hline
\multicolumn{3}{c}{HGS}                              & /             & 7 h                   & 0               & 0               & 0               & 0               & 0               & 0               & 0               & 0               & 0               & 0               & 0               & 0               \\
\multicolumn{3}{c}{LKH3}                             & /             & 7 h                   & 0.05\%          & 1.29\%          & 0.94\%          & 0.40\%          & 0.96\%          & /               & /               & /               & /               & /               & /               & 0.73\%          \\ \hline
\multirow{5}{*}{POMO}           & 1  & POMO\_CVRP      & 3.8 d         & \multirow{5}{*}{15 s} & \textbf{0.11\%} & 38.93\%         & 19.57\%         & 5.38\%          & 0.83\%          & 6.65\%          & 19.65\%         & 39.54\%         & 42.66\%         & 42.48\%         & 43.22\%         & 23.55\%         \\
  & 2  & POMO\_VRPTW     & 4.6 d         &                       & 29.34\%         & \textbf{2.52\%} & 66.48\%         & 58.91\%         & 28.06\%         & 53.55\%         & 66.66\%         & 1.60\%          & 7.70\%          & 7.91\%          & 8.00\%          & 30.06\%         \\
  & 3  & POMO\_OVRP      & 4.3 d         &                       & 9.28\%          & 45.48\%         & \textbf{2.11\%} & 15.79\%         & 9.23\%          & 17.12\%         & 2.38\%          & 44.65\%         & 26.80\%         & 27.32\%         & 27.16\%         & 20.67\%         \\
  & 4  & POMO\_VRPB      & 3.9 d         &                       & 1.85\%          & 42.61\%         & 16.64\%         & \textbf{2.10\%} & 1.90\%          & 3.05\%          & 16.75\%         & 43.90\%         & 42.60\%         & 41.55\%         & 41.77\%         & 23.16\%         \\
  & 5  & POMO\_VRPL      & 3.9 d         &                       & 0.49\%          & 38.73\%         & 19.70\%         & 5.48\%          & \textbf{0.57\%} & 6.07\%          & 19.44\%         & 39.62\%         & 42.34\%         & 42.41\%         & 43.49\%         & 23.48\%         \\ \hline
\multirow{12}{*}{Unified Model} & 6  & ST\_CVRP      & 4.1 d         & \multirow{7}{*}{20 s} & \textbf{0.52\%} & 39.11\%         & 19.52\%         & 5.37\%          & 0.59\%          & 7.07\%          & 19.02\%         & 41.10\%         & 42.34\%         & 41.64\%         & 41.88\%         & 23.47\%         \\
  & 7  & ST\_VRPTW     & 4.8 d         &                       & 39.15\%         & \textbf{2.16\%} & 66.23\%         & 67.85\%         & 37.40\%         & 55.30\%         & 65.62\%         & \textbf{1.28\%} & 7.83\%          & 7.99\%          & 7.91\%          & 32.61\%         \\
  & 8  & ST\_OVRP      & 4.4 d         &                       & 8.39\%          & 47.46\%         & \textbf{2.01\%} & 16.24\%         & 8.49\%          & 17.03\%         & \textbf{2.12\%} & 45.92\%         & 25.83\%         & 25.93\%         & 26.76\%         & 20.56\%         \\
  & 9  & ST\_VRPB      & 4.1 d         &                       & 1.22\%          & 41.50\%         & 16.94\%         & \textbf{1.80\%} & 1.59\%          & 3.46\%          & 17.32\%         & 44.67\%         & 43.50\%         & 42.06\%         & 41.40\%         & 23.23\%         \\
  & 10 & ST\_VRPL      & 4.3 d         &                       & 0.52\%          & 38.42\%         & 19.56\%         & 8.44\%          & \textbf{0.44\%} & 9.67\%          & 19.55\%         & 36.87\%         & 42.08\%         & 43.68\%         & 43.65\%         & 23.88\%         \\
  & 11 & ST\_OVRPBLTW  & 5.4 d         &                       & 22.95\%         & 7.40\%          & 50.29\%         & 43.11\%         & 20.77\%         & 33.84\%         & 48.63\%         & 6.62\%          & \textbf{0.90\%} & \textbf{0.84\%} & \textbf{0.86\%} & 21.47\%         \\
  & 12 & ST\_CVRP (FT) & 4.1 d + 0.6 d &                       & /          & 23.06\%         & 4.70\%          & 2.83\%          & 0.86\%          & 3.88\%          & 5.03\%          & 21.38\%         & 17.51\%         & 16.74\%         & 14.31\%         & 10.07\%         \\ \cline{2-17} 
  & 13 & MT (NN)       & 3.9 d         & \multirow{5}{*}{20 s} & 0.58\%          & 2.63\%          & 3.11\%          & 2.34\%          & 0.92\%          & 3.49\%          & 3.27\%          & 1.91\%          & 8.11\%          & 8.28\%          & 8.40\%          & 3.91\%          \\
  
  & 14 & MT (BN)       & 4.8 d         &                       & 0.55\%          & 2.66\%          & 3.29\%          & 2.47\%          & 1.11\%          & 3.43\%          & 3.27\%          & 1.82\%          & 7.82\%          & \textbf{7.88\%} & \textbf{7.82\%} & \textbf{3.83\%} \\
  
  & 15 & MT (IN)       & 4.8 d         &                       & 0.42\%          & 2.42\%          & 3.50\%          & 2.05\%          & 1.07\%          & \textbf{3.28\%} & \textbf{3.18\%} & \textbf{1.64\%} & \textbf{7.71\%} & 8.95\%          & 8.05\%          & \textbf{3.75\%} \\
  
  & 16 & MT (RN)       & 4.3 d         &                       & 0.55\%          & 2.50\%          & 3.34\%          & 2.24\%          & 0.93\%          & \textbf{3.31\%} & 3.35\%          & 2.19\%          & 7.82\%          & 8.08\%          & 8.13\%          & 3.86\%          \\
  & 17 & MT (IN, FT)   & 4.8 d + 0.6 d &                       & \textbf{0.37\%} & \textbf{2.42\%} & \textbf{3.04\%} & \textbf{2.05\%} & \textbf{0.86\%} & \textbf{3.15\%} & \textbf{3.06\%} & \textbf{1.63\%} & \textbf{1.90\%} & \textbf{1.84\%} & \textbf{1.96\%} & \textbf{2.02\%} \\ \hline
\end{tabular}%
}
\end{table}

\end{landscape}

\renewcommand{\arraystretch}{1.0}

\begin{table*}[htbp]
\centering

\caption{Experimental Results on Five New VRPs.}
\label{table:newvrps}

\resizebox{0.8\textwidth}{!}{%
\begin{tabular}{clccc|clccc}

\toprule
\multicolumn{1}{l}{Problem} & Method    & Dis.  & Gap     & Time & \multicolumn{1}{l}{Problem} & Method    & Dis.      & Gap     & Time \\

\midrule
\multirow{8}{*}{VRPBL}      & HGS& 11.15 & -& 14 h & \multirow{8}{*}{OVRPLTW}    & HGS& 17.35     & -& 14 h \\
   & NI & 17.38 & 55.92\% & 8m   &    & NI & 26.16     & 50.78\% & 8m   \\
   & FI & 16.37 & 46.88\% & 8m   &    & FI & 25.81     & 48.79\% & 8m   \\
   & POMO\_CVRP  & 12.41 & 11.37\% & 1.1m &    & POMO\_CVRP  & 26.75     & 54.18\% & 1.1m \\
   & POMO\_VRPTW & 17.86 & 60.27\% & 1.1m &    & POMO\_VRPTW & 19.41     & 11.88\% & 1.1m \\
   & Ours      & 11.65 & 4.50\%  & 1.2m &    & Ours      & 19.34     & 11.50\% & 1.2m \\
   & Ours FTD  & \textbf{11.62} & \textbf{4.27\%} & 1.2m &    & Ours FTD  & 18.12     & 4.46\%  & 1.2m \\
   & Ours FT   & 11.64 & 4.47\%  & 1.2m &    & Ours FT   & \textbf{17.86}    & \textbf{2.94\%} & 1.2m \\
   \midrule
\multirow{8}{*}{OVRPL}      & HGS& 9.71  & -& 14 h & \multirow{8}{*}{OVRPBTW}    & HGS& 17.31     & -& 14 h \\
   & NI & 13.61 & 40.17\% & 8m   &    & NI & 26.24     & 51.60\% & 8m   \\
   & FI & 13.46 & 38.62\% & 8m   &    & FI & 25.79     & 48.99\% & 8m   \\
   & POMO\_CVRP  & 11.78 & 21.35\% & 1.1m &    & POMO\_CVRP  & 26.78     & 54.69\% & 1.1m \\
   & POMO\_VRPTW & 18.62 & 91.78\% & 1.1m &    & POMO\_VRPTW & 19.48     & 12.51\% & 1.1m \\
   & Ours      & 10.15 & 4.57\%  & 1.2m &    & Ours      & 19.32     & 11.61\% & 1.2m \\
   & Ours FTD  & 10.18 & 4.87\%  & 1.2m &    & Ours FTD  & 18.12     & 4.70\%  & 1.2m \\
   & Ours FT   & \textbf{10.15} & \textbf{4.57\%} & 1.2m &    & Ours FT   & \textbf{17.84 }& \textbf{3.07\%}  & 1.2m \\
   \midrule
\multirow{8}{*}{VRPBTW}     & HGS& 26.31 & -& 14 h & \multirow{8}{*}{Average}    & HGS& 16.36     & -& 14 h \\
   & NI & 36.84 & 40.05\% & 8m   &    & NI & 24.05     & 46.95\% & 8m   \\
   & FI & 36.59 & 39.10\% & 8m   &    & FI & 23.60     & 44.25\% & 8m   \\
   & POMO\_CVRP  & 38.99 & 48.20\% & 1.1m &    & POMO\_CVRP  & 23.34     & 42.64\% & 1.1m \\
   & POMO\_VRPTW & 27.18 & 3.33\%  & 1.1m &    & POMO\_VRPTW & 20.51     & 25.34\% & 1.1m \\
   & Ours      & 27.11 & 3.05\%  & 1.2m &    & Ours      & 17.51     & 7.03\%  & 1.2m \\
   & Ours FTD  &\textbf{27.08}& \textbf{2.94\%} & 1.2m &    & Ours FTD  & 17.03     & 4.05\%  & 1.2m \\
   & Ours FT   & 27.11 & 3.05\%  & 1.2m &    & Ours FT   & \textbf{16.92}   & \textbf{3.40\%}  & 1.2m \\
   \bottomrule
\end{tabular}%
}
\end{table*}

The results indicate that fine-tuning can further improve the performance of our model on new VRPs. Our pre-trained model with fast fine-tuning outperforms all other methods (except for the baseline). The average gap to the baseline HGS is only about 3.4\% over the five VRPs. The advantages of fast fine-tuning become more apparent on OVRPLTW and OVRPBTW, which have more attributes and are therefore more complicated. In contrast, fine-tuning only provides minor improvements on VRPBL, OVRPL, and VRPBTW, where zero-shot generalization has already produced satisfactory results.

\begin{figure}[htbp]
    \centering
    \includegraphics[width=0.5\textwidth]{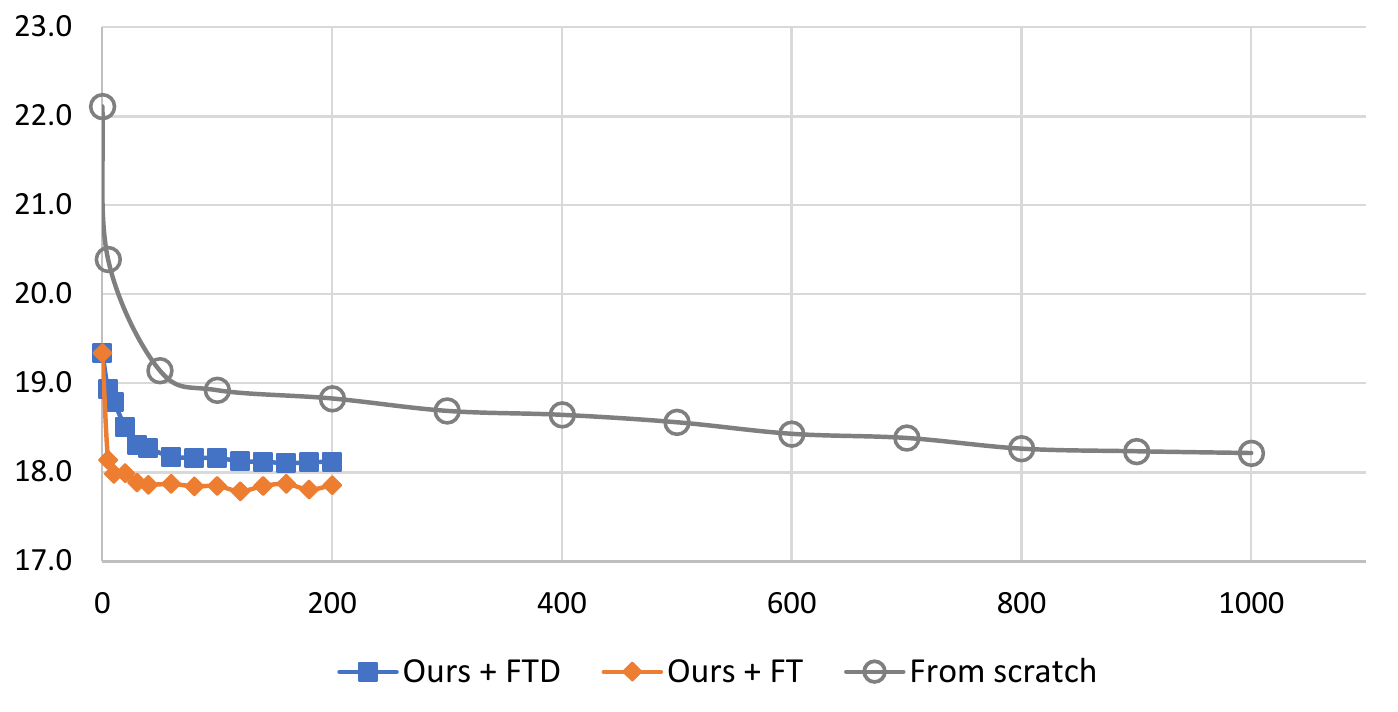}
    \caption{Testing distance vs. epoch number. FTD and FT indicate fine-tuning on the decoder and the entire network of our pre-train unified model, respectively. The curve with grey circles represents training a model from scratch. Each epoch includes 10,000 randomly generated OVRPLTW instances. The testing distance is the average result over 2,000 OVRPLTW instances.}
    \label{fig:fine_tune}
\end{figure}

Figure~\ref{fig:fine_tune} illustrates the convergence of testing distance vs. the number of fine-tuning epochs on OVRPLTW. We compare the performance of fine-tuning the pre-trained model with training a new model from scratch. The experimental settings of training from scratch are the same as that used for fine-tuning. The results demonstrate that fine-tuning based on our pre-trained unified model converges rapidly. The fine-tuning of the entire pre-trained model takes less than 50 epochs (about 1.5 hours) to converge, outperforming the results of training from scratch using 1,000 epochs. The advantage of fast adaptation highlights the significance of our pre-trained unified model.

The results demonstrate the promising generalization ability of our unified model on new problems. In addition to achieving overall acceptable results through direct zero-shot generalization, our proposed unified model can be fine-tuned on new problems with low computational cost, further improving the solution quality.




\section{Experiments on Out-of-Distribution Instances}~\label{sec:d}

Existing works on neural combinatorial optimization typically assume that the training and testing instances come from the same distribution. Achieving generalization to out-of-distribution cases is often challenging for traditional single-task models. In this section, we investigate the generalization performance of our proposed multi-task model on three out-of-distribution scenarios of the training problems. All of our experiments are carried out on the problems of size $100$.


\subsection{CVRP} 

\textbf{Basic Settings:} The training instance generation for CVRP follows the distribution used in~\citet{kool2018attention}. The coordinates of customers $x_i,y_i$ are randomly sampled in the unite region $[0,1]$. The demands of customers $d_i$ are randomly selected from $\{1,\dots,9\}$ and then normalized with respect to the vehicle capacity $C$. The capacity is set to be $C=40$ and $C=50$ for the problem with a size of $n=50$ and $n=100$, respectively.

\textbf{Out-of-Distribution Settings:}We evaluate the performance of our model with POMO on CVRP instances with different vehicle capacities. Specifically, we consider seven vehicle capacities: $C=\{20, 30, 40, 50, 60, 70, 80\}$.

\begin{figure}[htbp]
    \centering
    \includegraphics[width=0.5\textwidth]{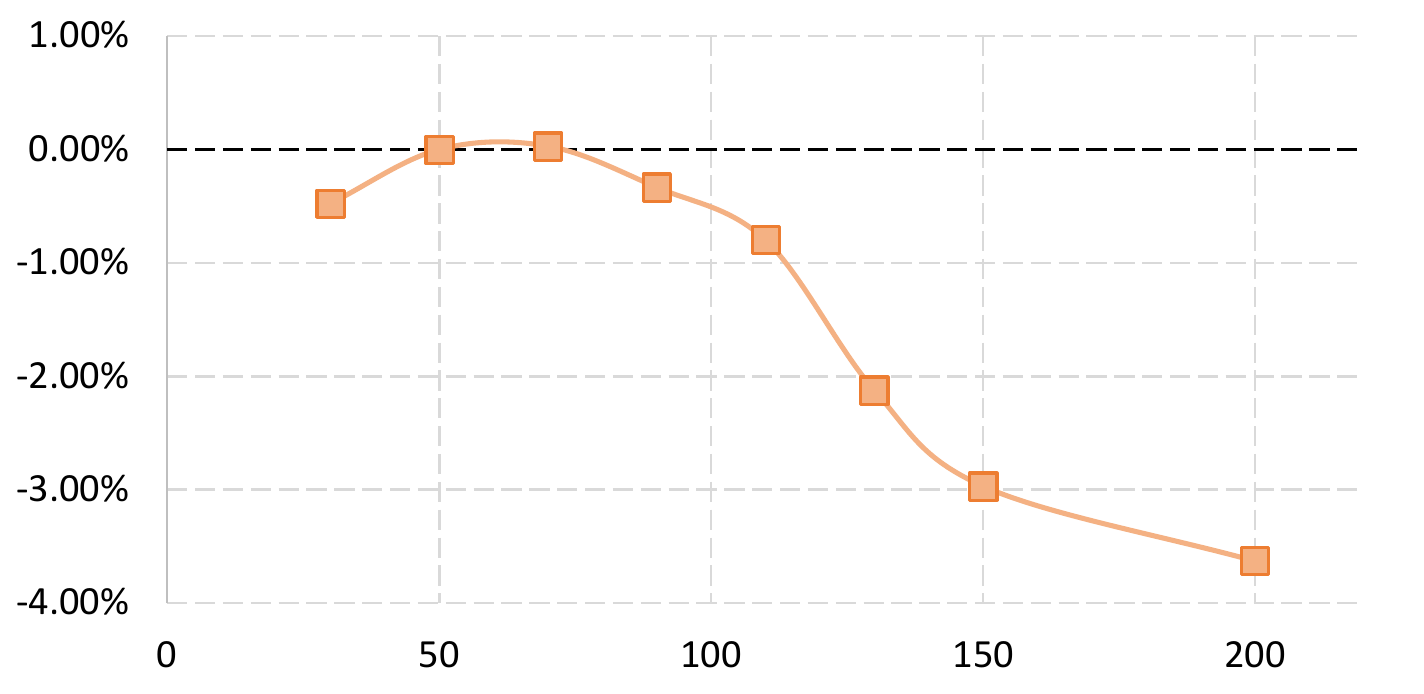}
    \caption{Performance gap between Ours and POMO with respect to different vehicle capacities on CVRP.}
    \label{fig:outcvrp}
\end{figure}

Table~\ref{table:outcvrp} and Figure~\ref{fig:outcvrp} show the results on out-of-distribution CVRP instances averaged over 5,000 instances. The better results are in bold, and the gap is calculated using POMO as the baseline. A small gap is preferred and a negative gap indicates that our model outperforms POMO.  Except for the results on 50 and 70 (close to that used for generating training instances), our model beats POMO on all the out-of-distribution cases. The advantage of using our unified model is more obvious with the increasing vehicle capacity.

\begin{table}[htbp]
\centering
\caption{A comparison of our model with single-task POMO on out-of-distribution CVRP instances}
\label{table:outcvrp}
\resizebox{0.5\textwidth}{!}{%
\begin{tabular}{ccccccccc}
\toprule
Vehicle capacity   & 30  & 50  & 70  & 90  & 110 & 130& 150& 200\\
\midrule

POMO\_CVRP & 22.913   & \textbf{15.750} & \textbf{12.910} & 11.480   & 10.595   & 10.114  & 9.824   & 9.307   \\

Ours& \textbf{22.804} & 15.750   & 12.914   & \textbf{11.441} & \textbf{10.511} & \textbf{9.900} & \textbf{9.532} & \textbf{8.969} \\
\midrule

Gap & -0.48\%  & 0.00\%   & 0.03\%   & -0.33\%  & -0.80\%  & -2.12\% & -2.97\% & -3.63\%\\

\bottomrule
\end{tabular}%
}
\end{table}

\subsection{VRPTW}

\textbf{Basic Settings:} For VRPTW, we adopt the same settings as those used in~\citet{zhao2020hybrid}. Specifically, we generate the coordinates, demands, and vehicle capacities using the same procedure as for CVRP. The time windows attributes involve three additional features: 1) the service time $s_i$, 2) the early time $e_i$, and 3) the late time $l_i$. The service time $s_i$ and the length of the time window $\Delta_i$ are randomly sampled from a closed interval $[0.15,0.2]$. The vehicle speed is fixed at $v=1$ and the maximum time interval of the depot is set to $T=4.6$. It means that all vehicles must return to the depot before the maximum time interval.

The early and late times for each customer can be calculated as follows:
\begin{equation}
    \begin{aligned}
 e_i & = \frac{h_i \times c_{0 i}}{v}, 
 h_i \in\left[1, \frac{T-s_i-\Delta_i}{c_{0 i}} \times v-1\right] \\
 l_i & = e_i+\Delta_i
    \end{aligned}
\end{equation}
where $c_{0i}$ denotes the distance between the depot and the $i$-th customer. This formulation ensures that there is at least one feasible solution for each instance.

\textbf{Out-of-Distribution Settings:} For out-of-distribution testing, we only modify the closed time interval used for sampling service time and time window length. The intervals are as follows: $\{[0.05,0.1],[0.15,0.2], \dots, [0.85,0.9], [0.95,1.0]\} $.

\begin{table}[t]
\centering
\caption{A comparison of our model with single-task POMO on out-of-distribution VRPTW instances. }
\label{table:outvrptw}
\resizebox{0.5\textwidth}{!}{%
\begin{tabular}{cccccc}
\toprule
Time   Interval & \multicolumn{1}{c}{{[}0.05,0.1{]}}  & 
\multicolumn{1}{c}{{[}0.15,0.2{]}} & \multicolumn{1}{c}{{[}0.25,0.3{]}}  & \multicolumn{1}{c}{{[}0.35,0.4{]}}  & \multicolumn{1}{c}{{[}0.45,0.5{]}}  \\
\midrule
POMO\_VRPTW     & \multicolumn{1}{c}{\textbf{26.230}}          & \multicolumn{1}{c}{\textbf{26.906}}                        & \multicolumn{1}{c}{\textbf{28.805}} & \multicolumn{1}{c}{32.327}          & \multicolumn{1}{c}{36.544}          \\
Ours            & \multicolumn{1}{c}{26.390} & \multicolumn{1}{c}{27.152}     & \multicolumn{1}{c}{29.046}          & \multicolumn{1}{c}{\textbf{32.174}} & \multicolumn{1}{c}{\textbf{36.162}} \\
\midrule
Gap             & \multicolumn{1}{c}{0.61\%}          & \multicolumn{1}{c}{0.91\%}     & \multicolumn{1}{c}{0.84\%}          & \multicolumn{1}{c}{-0.47\%}         & \multicolumn{1}{c}{-1.05\%}         \\
\midrule
\midrule
Time Interval   & \multicolumn{1}{c}{{[}0.55,0.6{]}} & \multicolumn{1}{c}{{[}0.65,0.7{]}}     & \multicolumn{1}{c}{{[}0.75,0.8{]}} & \multicolumn{1}{c}{{[}0.85,0.9{]}} & \multicolumn{1}{c}{{[}0.95,1.0{]}} \\
\midrule
POMO\_VRPTW     & 40.650 & 45.327                       & 49.219 & 53.300              & 57.102 \\
Ours            & \textbf{40.136}                    & \textbf{45.007}     & \textbf{48.901}                    & \textbf{52.815} & \textbf{56.209}                    \\
\midrule
Gap             & -1.27\%& -0.71\%    & -0.65\%& -0.91\%& -1.56\%      \\
\bottomrule
\end{tabular}%
}
\end{table}

\begin{figure}[t]
    \centering
    \includegraphics[width=0.5\textwidth]{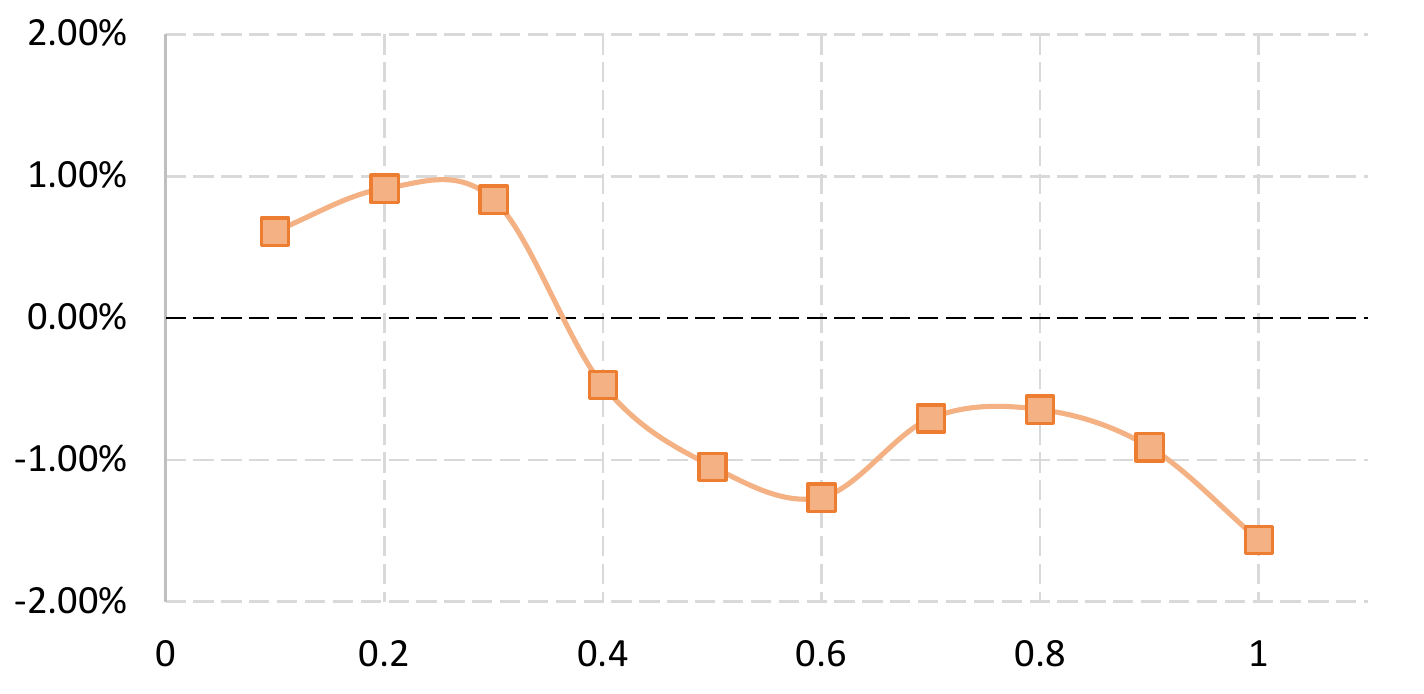}
    \caption{Performance gap between Ours and POMO with respect to different time intervals on VRPTW.}
    \label{fig:outvrptw}
\end{figure}

The results in Table~\ref{table:outvrptw} and Figure~\ref{fig:outvrptw} again reveal that our model is more robust in out-of-distribution cases. We have negative gaps in most cases except for these near training distribution $[0.15,0.2]$.

\subsection{VRPL}

\textbf{Basic Settings:} Same as CVRP.

\textbf{Out-of-distribution Settings:} In the training, the duration limit is fixed to $l=3.0$. We modify the duration limit of each route. The settings used are $l=\{2.9,3.0,3.1,3.2,3.3,3.4,3.5\}$. To better show the effect of the duration limit on the results, we increase the vehicle capacity from 50 to 150 for VRPL. In this way, the number of nodes in each route will increase and the influence of duration limit will be more significant. 

\begin{figure}[t]
    \centering
    \includegraphics[width=0.5\textwidth]{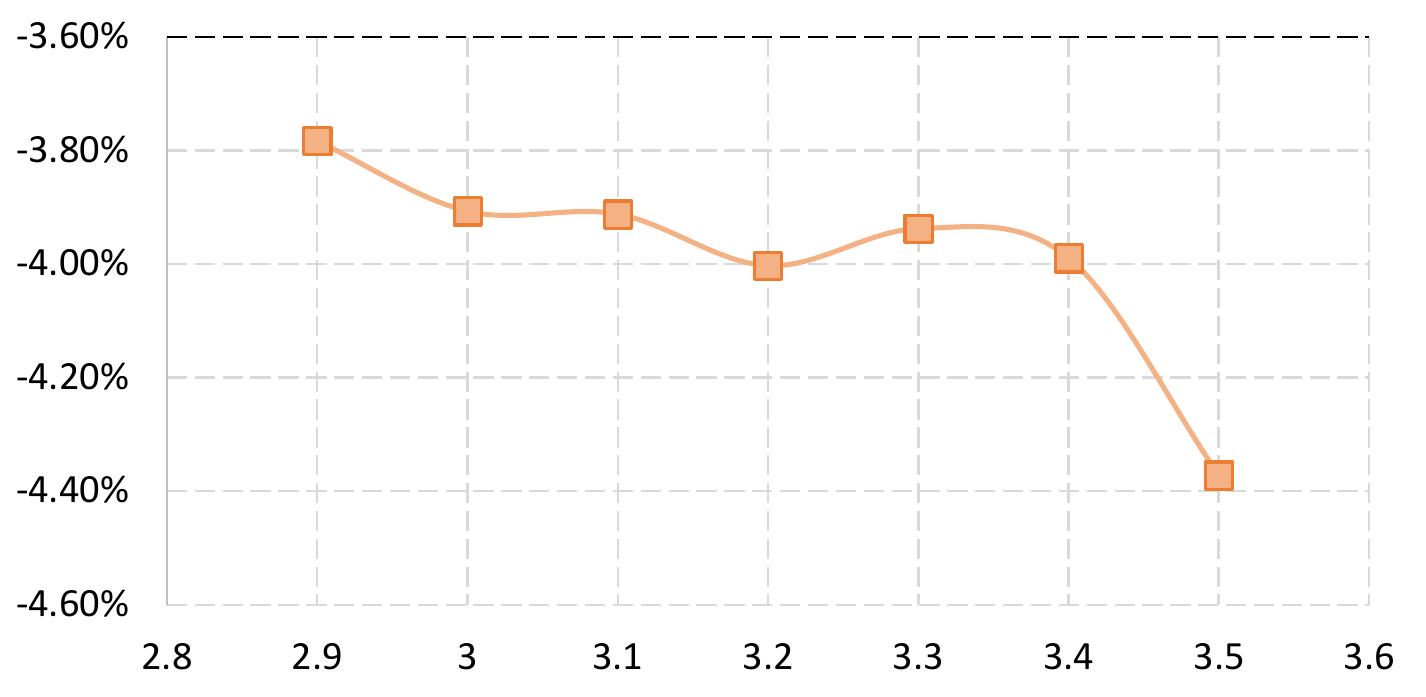}
    \caption{Performance gap between Ours and POMO with respect to different duration limitations on VRPL.}
    \label{fig:outvrpl}
\end{figure}

Table~\ref{table:outvrpl} and Figure~\ref{fig:outvrpl} show the results. Our model outperforms POMO in all distributions (the reason might be that we have also modified the vehicle capacity). The advantage of our model is more obvious on large duration limits.

\begin{table}[t]
\centering
\caption{A comparison of our model with single-task POMO on out-of-distribution VRPL instances. }
\label{table:outvrpl}
\resizebox{0.5\textwidth}{!}{%
\begin{tabular}{cccccccc}
\toprule
Duration   & 2.9            & 3 & 3.1             & 3.2            & 3.3            & 3.4            & 3.5            \\
\midrule
POMO\_VRPL & 10.245         & 10.145          & 10.077 & 10.044         & 10.019         & 10.000         & 9.998          \\
Ours       & \textbf{9.858} & \textbf{9.748 }                    & \textbf{9.683}           & \textbf{9.642} & \textbf{9.625} & \textbf{9.601} & \textbf{9.561} \\
\midrule
Gap        & -3.78\%        & -3.91\%                   & -3.91\%         & -4.00\%        & -3.94\%        & -3.99\%        & -4.37\%     \\
\bottomrule
\end{tabular}%
}
\end{table}

Our experiments reveal that our multi-task model displays a strong generalization performance across various distributions. In contrast, single-task models perform best only on the test scenarios that share the same distributions as the training data. They are outperformed by the multi-task model when it comes to out-of-distribution scenarios. 

We observed that a multi-task model's advantage becomes more pronounced in extreme cases. For example, on CVRP with a vehicle capacity of 200, our model exhibits a substantial performance gap compared to POMO models. One possible explanation for this is that training a multi-task model with different VRP variants brings diversity. Although it does not specifically represent the utilization of various vehicle capacities, the multi-task model can extract more comprehensive patterns than single-task learning.

\end{document}